%% file: main.tex
\documentclass[oneside,article]{memoir}

\usepackage{amsmath}         
\usepackage{amsthm,thmtools,thm-restate} 
\usepackage{enumitem}        

\usepackage{rotating}

\usepackage[final]{listings}
\usepackage{bbm}
\usepackage{xcolor}

\definecolor{codegreen}{rgb}{0,0.6,0}
\definecolor{codegray}{rgb}{0.5,0.5,0.5}
\definecolor{codepurple}{rgb}{0.58,0,0.82}
\definecolor{backcolour}{rgb}{0.95,0.95,0.92}

\lstdefinestyle{mystyle}{
    backgroundcolor=\color{backcolour},   
    commentstyle=\color{codegreen},
    keywordstyle=\color{magenta},
    numberstyle=\tiny\color{codegray},
    stringstyle=\color{codepurple},
    basicstyle=\ttfamily\footnotesize,
    breakatwhitespace=false,         
    breaklines=true,                 
    captionpos=b,                    
    keepspaces=true,                 
    numbers=left,                    
    numbersep=5pt,                  
    showspaces=false,                
    showstringspaces=false,
    showtabs=false,                  
    tabsize=2
}

\usepackage{algorithm}
\usepackage{algpseudocode}
\usepackage{algorithmicx}
\usepackage{threeparttable}
\usepackage{booktabs}
\usepackage{float}

\makeatletter
\providecommand{\ALG@name}{Algorithm}   

\makeatother

\providecommand{\needspace}[1]{}

\usepackage{tikz}
\usetikzlibrary{matrix,arrows,positioning}

\usepackage{pgfplots}
\pgfplotsset{compat=1.17}
\usepgfplotslibrary{groupplots}

\definecolor{dhcol}{RGB}{45,110,165}   
\definecolor{vrcol}{RGB}{215,140,40}   
\definecolor{qcol}{RGB}{170,60,50}     
\definecolor{acol}{RGB}{40,130,90}     

\pgfplotsset{
  redzedplot/.style={
    width=0.47\textwidth, height=5.2cm,
    ymajorgrids, grid style={gray!25},
    label style={font=\small}, tick label style={font=\footnotesize},
    title style={font=\small}, legend style={font=\footnotesize, draw=none,
      fill=white, fill opacity=0.85, text opacity=1, cells={anchor=west}},
    every axis plot/.append style={thick},
    log basis y=10,
  },
}

\tikzset{
  dhline/.style={dhcol, mark=*, mark size=1.6pt},
  vrline/.style={vrcol, mark=square*, mark size=1.6pt},
  qline/.style={qcol, mark=triangle*, mark size=2.2pt},
  estline/.style={dashed, mark options={fill=white}},
}
\usepackage{amsfonts}
\usepackage{parskip}
\usepackage{mathtools}
\usepackage{wrapfig}

\usepackage{tikz-cd}

\usepackage{caption, subcaption} 

\usepackage[hidelinks]{hyperref}                   
\usepackage[capitalize,nosort]{cleveref}           

\usepackage{amssymb}
\usepackage[mathscr]{euscript}
\usepackage{relsize}
\usepackage{stmaryrd}
\usepackage{stackengine}

\usepackage{indentfirst} 
\usepackage{multicol}

\usepackage{graphicx}
\usepackage{ragged2e} 
\usepackage[inline,nomargin,draft]{fixme}

\isopage[12]

\setlrmargins{*}{*}{1}
\checkandfixthelayout

\counterwithout{section}{chapter}
\usepackage{appendix}

\makeatletter
\providecommand{\abx@aux@refcontext}[1]{}
\providecommand{\abx@aux@cite}[2]{}
\providecommand{\abx@aux@segm}[3]{}
\providecommand{\abx@aux@page}[2]{}
\providecommand{\abx@aux@fnpage}[2]{}
\providecommand{\abx@aux@backref}[5]{}
\providecommand{\abx@aux@defaultrefcontext}[3]{}
\providecommand{\abx@aux@read@bbl@mdfivesum}[1]{}
\providecommand{\abx@aux@read@bblrerun}{}
\providecommand{\abx@aux@sortscheme}[1]{}
\providecommand{\abx@aux@refsection}[1]{}
\providecommand{\abx@aux@number}[2]{}
\makeatother

\makeatletter
\let\dgm\@undefined
\makeatother

\input{defs}

\tikzstyle{vertex}=[circle, draw, minimum size=7pt, inner sep=0pt]

\usepackage[T1]{fontenc}
\usepackage{amssymb}
\usepackage{array} 
\usepackage{longtable}
\usepackage{booktabs}
\usepackage{hyperref} 

\DeclareFontFamily{U}{dmjhira}{}
\DeclareFontShape{U}{dmjhira}{m}{n}{ <-> dmjhira }{}

\author{Krzysztof Kapulkin \thanks{College of Connected Computing \& Department of Mathematics, Vanderbilt University} \and 
Nathan Kershaw \thanks{Department of Mathematics, University of Western Ontario}}

\title{Reasoning is topologically simple}

\date{\today}

\begin{document}

  \maketitle

\begin{abstract}
  We formulate and investigate the hypothesis that reasoning text has simpler co-occurrence topology than non-reasoning text, and that higher-quality reasoning has simpler co-occurrence topology than lower-quality reasoning. 
  We discuss and test applications to AI safety and AI evaluation. 
\end{abstract}



\input{0-introduction.tex}
\input{1-methodology.tex}

\input{2-results.tex}
\input{3-discussion.tex}

\bibliographystyle{amsalphaurlmod}
\bibliography{all-refs}

 \appendix
 \renewcommand{\thesection}{\Alph{section}}
 \begin{appendices}
 \input{A-cooccurrence.tex}
 \input{B-experiments}
 \end{appendices}

\end{document}

%% file: defs.tex
  \renewcommand{\hat}{\widehat}

  \declaretheorem[style=definition,within=section]{definition}

  \declaretheorem[style=definition,numberlike=definition]{hypothesis}

  \declaretheorem[style=plain,numberlike=definition]{proposition}
  \declaretheorem[style=plain,numberlike=definition]{theorem}

  \declaretheorem[style=plain,numbered=no,name=Theorem]{theorem*}

  \Crefname{corollary}{Corollary}{Corollaries}
  \Crefname{definition}{Definition}{Definitions}
  \Crefname{lemma}{Lemma}{Lemmas}
  \Crefname{proposition}{Proposition}{Propositions}
  \Crefname{remark}{Remark}{Remarks}
  \Crefname{theorem}{Theorem}{Theorems}
  \Crefname{notation}{Notation}{Notations}
  \Crefname{conjecture}{Conjecture}{Conjectures}
    \crefname{hypothesis}{Hypothesis}{Hypotheses}

  \newlist{axioms}{enumerate}{1}
  \Crefname{axiomsi}{}{}

  \newenvironment{tikzeq*}
  {
    \begingroup
    \begin{equation*}
    \begin{tikzpicture}[baseline=(current bounding box.center)]
  }
  {
    \end{tikzpicture}
    \end{equation*}
    \endgroup
    \ignorespacesafterend
  }

  \tikzset
  {
    diagram/.style=
    {
      matrix of math nodes,
      column sep={4.3em,between origins},
      row sep={4em,between origins},
      text height=1.5ex,
      text depth=.25ex
    },
    over/.style={preaction={draw=white,-,line width=6pt}},
    every to/.style={font=\footnotesize},
    inj/.style={right hook->},
    surj/.style={-{Latex[open]}},
    cof/.style={>->},
    fib/.style={->>},
  }

  \DeclareFontFamily{U}{mathx}{\hyphenchar\font45}

  \DeclareFontShape{U}{mathx}{m}{n}{
    <5> <6> <7> <8> <9> <10>
    <10.95> <12> <14.4> <17.28> <20.74> <24.88>
    mathx10}{}

  \DeclareSymbolFont{mathx}{U}{mathx}{m}{n}

  \DeclareFontFamily{U}{mathb}{\hyphenchar\font45}

  \DeclareFontShape{U}{mathb}{m}{n}{
    <5> <6> <7> <8> <9> <10>
    <10.95> <12> <14.4> <17.28> <20.74> <24.88>
    mathb10}{}

  \DeclareSymbolFont{mathb}{U}{mathb}{m}{n}

  \DeclareMathAccent{\widebar}{0}{mathx}{"73}

  \DeclareMathSymbol{\Rsh}{\mathrel}{mathb}{"E9}

  \DeclareFontFamily{U}{MnSymbolA}{}

  \DeclareFontShape{U}{MnSymbolA}{m}{n}{
    <-6> MnSymbolA5
    <6-7> MnSymbolA6
    <7-8> MnSymbolA7
    <8-9> MnSymbolA8
    <9-10> MnSymbolA9
    <10-12> MnSymbolA10
    <12-> MnSymbolA12}{}

  \DeclareSymbolFont{MnSyA}{U}{MnSymbolA}{m}{n}

  \DeclareMathSymbol{\twoheaddownarrow}{\mathrel}{MnSyA}{27}

  \newcommand{\MSC}[1]{%
    \let\thempfn\relax
    \footnotetext[0]{2020 Mathematics Subject Classification: #1.}
  }

  \newcommand{\dgm}{\textsf{dgm}}

  \renewcommand{\dgm}{\operatorname{dgm}}

%% file: 0-introduction.tex
\section*{Introduction}

Since the introduction of chain-of-thought (CoT) prompting \cite{wei2022chain,kojima2022zeroshot} in large language models (LLMs), it is commonplace for LLMs to reason in long natural-language traces.
These traces are used both for evaluating models \cite{golovneva2023roscoe,prasad2023receval,lightman2024verify,zheng2025processbench} and for AI safety \cite{baker2025monitoring,korbak2025monitorability,guan2026monitoring,kutasov2025shade}.
A common approach to evaluating these traces is to use a separate LLM as a monitor; however, this LLM monitor is expensive, shares biases with the LLM it is judging, and may be subject to the same safety concerns as the original LLM.
Developing methods and statistics to evaluate reasoning traces is thus an important problem in both AI evaluation and safety.
Such methods have the added benefit of being able to evaluate other sorts of writing, such as human writing or AI-written papers.

Topological data analysis (TDA) has been proposed as a tool to evaluate the reasoning of LLMs \cite{tan2026shape,li2026understanding,zhang2026global,zhang2026tdarc,more2025optimizing}.
A common approach is to fix an embedding of reasoning steps (sentences, ideas, etc.) into $\mathbb{R}^n$, and compute Vietoris--Rips persistent homology.
While this approach has demonstrated success, it also comes with drawbacks: the order of the reasoning steps can be lost, or the influence muted; it relies on choices for the embedding; and the answers are often hard to interpret (i.e. there often lacks an interpretation of what a hole represents in terms of reasoning steps and ideas).
A natural solution to these issues is to use co-occurrence persistent homology instead:
Each sentence represents a simplex added into the filtration, and the vertices are the words in the sentence.
This has the benefit that it uses reasoning order directly for the filtration order, requires few choices, and gives interpretable results (a 1-dimensional hole, for example, represents a cycle of words used pairwise but not yet combined into one idea).
Using co-occurrence persistent homology to evaluate text is not a novel idea and has been used before \cite{salnikov2018cooccurrence,christianson2020architecture}.
To our knowledge, however, this construction has not yet been used to evaluate reasoning quality directly. 

When considering co-occurrence persistence to evaluate reasoning, there is a natural hypothesis.
When reasoning, concepts are not only introduced, but combined and used together.
If a collection of three or more concepts is used pairwise together, one would expect them to often be further combined and used together.
This would correspond to the creation and subsequent filling in of a hole (see \cref{fig:cooccurrence}).
Thus, a natural hypothesis is that reasoning has simpler topology than non-reasoning, and higher-quality reasoning has simpler topology than lower-quality reasoning.
To measure topological complexity, we compute persistent homology in dimension 1.
The \textit{total persistence} is defined as the sum of lengths of all intervals in the persistence diagram, which we then normalize by dividing by the mean total persistence when randomly shuffling the text.
Any number smaller than 1 indicates simpler topology than random, with even smaller numbers representing even simpler topology.

It is important to distinguish this explanatory hypothesis from the empirical claims tested below. Our experiments compare co-occurrence topology across text genres, human- and AI-written papers, models of differing capability, and benign and sabotage conditions. We propose that the relationship between reasoning and topological simplicity provides a possible explanation for the resulting patterns, rather than assuming that these comparisons directly measure reasoning quality. The proposed relationship is a statistical tendency rather than a criterion for the quality or correctness of an individual text.

We run four experiments to test this idea and identify applications to AI evaluation and safety.
The first test is to compare reasoning prose with non-reasoning prose.
We have two reasoning categories and two non-reasoning categories: mathematics and argumentative philosophy texts for reasoning, and history and short stories for non-reasoning.
The second test compares 135 of the recent OpenAI mathematics papers \cite{openai2026math} to human-written counterparts.
LLM-written papers are often criticized as difficult to read, and the hypothesis here is that human papers often present a more refined, easy-to-follow form of reasoning, resulting in simpler topology.
Again, it is important to distinguish this explanatory hypothesis from what is actually being tested, as there are significant confounding variables such as writing style.
As a third test, we compare LLM capability with reasoning topology.
In particular, for 48 open-weight models, we compute co-occurrence persistence on reasoning traces for 30 AIME problems \cite{matharena_aime2026}, and compare each model's median with its Epoch Capabilities Index (ECI) \cite{EpochEci2025,ho2025rosetta} to test whether more capable models (higher-quality reasoning) have simpler topology.
Finally, we use the construction to evaluate deepseek-v4-flash-0731 on SHADE-arena tasks \cite{kutasov2025shade}, which test a model performing tasks under normal circumstances to a model given a normal task and a sabotage side task.
The hypothesis is that a model needing to balance a normal and sabotage task degrades its reasoning in a detectable manner.

The results are as follows: For the first experiment, we find that mathematics texts have simpler topology than both history ($p < 10^{-5}$) and short stories ($p < 10^{-5}$), and argumentative philosophy texts have simpler topology than history ($p=0.019$) and short stories ($p < 10^{-5}$).
For the second experiment, the human-written counterparts had simpler topology in around 93\% of pairs ($p=5 \times 10^{-23}$).
For the third experiment, ECI correlated negatively with average topology, having a Spearman correlation coefficient of $\rho = -0.444$ ($p=0.0008$).
Finally, for the SHADE-arena, the non-sabotage reasoning had simpler topology in 19/26 environments ($p=0.009$).
Altogether, these experiments provide strong support for the initial hypothesis and establish applications to both AI safety and evaluation.

All code and experimental results can be found at 
\begin{center}
    \url{https://github.com/nkershaw01/Co-occurrence-reasoning-topology}
\end{center}

\paragraph{Related work.}
There is a large body of work using TDA methods to analyze LLMs. 
The most prominent approach is to apply persistent homology to the activation space \cite{ruppik2024local,gardinazzi2025persistent,fay2026shape,pan2026topology}, however, there is existing work applying TDA constructions to reasoning traces as well. 
In \cite{tan2026shape}, the authors embed reasoning steps from AIME problems into $\mathbb{R}^n$ and use persistent homology derived metrics to predict reasoning quality. 
Similarly, Li et al.~\cite{li2026understanding} embed steps from chain-of-thought, tree-of-thought, and graph-of-thought reasoning, augment the embeddings with positional or structural encodings, and compute Vietoris--Rips persistent homology to compare their topological structure.
In particular, their treatment of step order is through the embedding rather than through the filtration parameter.
More et al.~\cite{more2025optimizing} instead embed multiple complete reasoning traces generated for the same prompt and combine topological features of this collection with Dirichlet-based uncertainty estimates to calibrate model confidence.
The novelty of our approach is to avoid the use of any embedding, avoiding Vietoris--Rips persistence altogether, and to use the natural ordering of reasoning steps as the filtration order. 
This provides interpretable results that are also independent of choices of embedding. 

Co-occurrence persistent homology has been used to analyze text. 
In \cite{salnikov2018cooccurrence}, the authors apply a co-occurrence construction to mathematics papers, and suggest that persistent holes correspond to gaps in knowledge. 
In \cite{christianson2020architecture} the authors apply a similar construction to evaluate mathematics textbooks, comparing topology to textbook ratings. Here, vertices are extracted concepts, and edges record sentence-level co-occurrence; the filtration follows sentence order and takes the clique complex at each stage.
The authors also suggest that persistent holes correspond to longstanding gaps in knowledge. 
To our knowledge, co-occurrence persistent homology has not yet been used to directly evaluate reasoning.

%% file: 1-methodology.tex
\section{Methodology} \label{sec:methodology}
In this section, we discuss the methodology of each of the four experiments, as well as the shared construction.

\subsection{Co-occurrence persistent homology}

We begin by discussing co-occurrence persistent homology, the tool underlying our approach.
We omit the full mathematical details in this section, aiming to provide enough for an unfamiliar reader to understand the broad idea behind the construction. 
For full mathematical details, see Appendix A.
The object of interest that we wish to study is text.
To do this, we read a text one sentence at a time and record which concepts (words) have been used together in a common sentence. 
Each sentence represents a set of words, and a text is therefore represented by a sequence of such sets. 

We can further represent this using a graph that grows as sentences are read. 
The vertices of the graph are words that have been seen so far, and an edge is drawn between two words if they appear in a common sentence.
When a new sentence is read, unused words are added to the graph as vertices, and new pairs of words appearing in the same sentence are added as edges. 
Observing how the graph changes over ``time" as sentences are read gives a \textit{filtration} of graphs. 

The problem that arises using graphs in this manner is that it only detects pairwise relations through edges, and useful higher-order information may be lost.
To solve this issue, we move to the setting of \textit{simplicial complexes}. 
We maintain the same underlying filtration of graphs, but glue in a triangle between three words whenever they all appear in a sentence together. Likewise, we glue in a tetrahedron between four words appearing together, and in general and $n$-simplex between $n+1$ words appearing in a sentence together. 
Tracking how this simplicial complex changes as words are read gives a filtration of simplicial complexes.

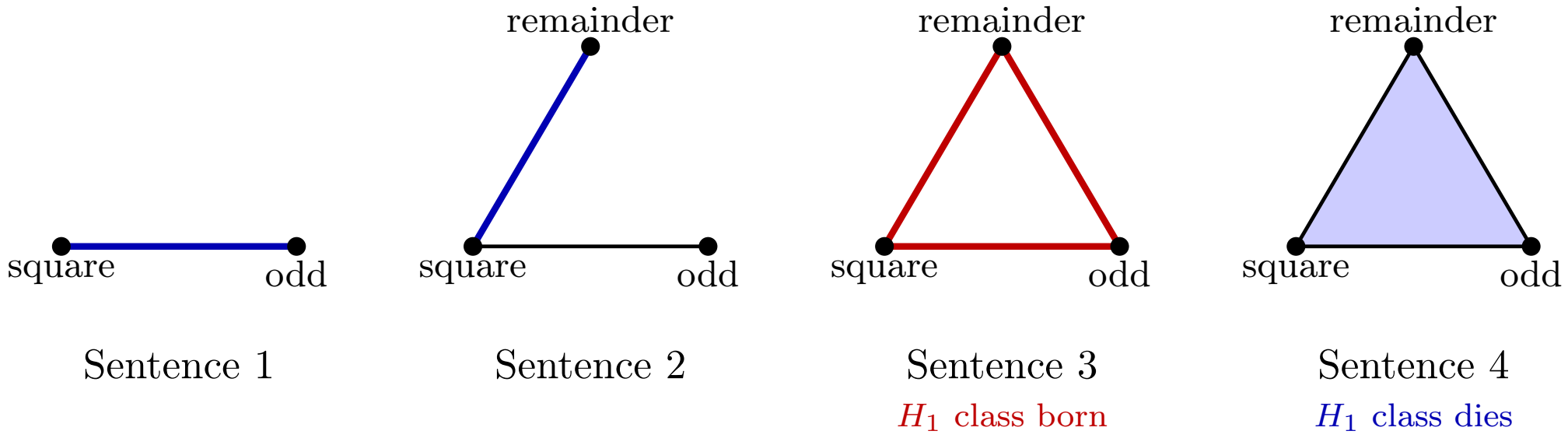
\begin{figure}[H] 
\centering
\begin{tikzpicture}[
    v/.style   = {circle, fill=black, inner sep=1.6pt},
    lab/.style = {font=\small},
    old/.style = {line width=0.9pt, black},
    new/.style = {line width=1.6pt, blue!70!black},
    cyc/.style = {line width=1.6pt, red!75!black},
  ]

  \begin{scope}[xshift=0cm]
    \coordinate (S) at (0,0); \coordinate (O) at (2,0);
    \draw[new] (S) -- (O);
    \node[v] at (S) {}; \node[v] at (O) {};
    \node[lab, below] at (S) {square};
    \node[lab, below] at (O) {odd};
    \node at (1,-1.0) {Sentence 1};
  \end{scope}

  \begin{scope}[xshift=3.5cm]
    \coordinate (S) at (0,0); \coordinate (O) at (2,0); \coordinate (R) at (1,1.7);
    \draw[old] (S) -- (O);
    \draw[new] (S) -- (R);
    \node[v] at (S) {}; \node[v] at (O) {}; \node[v] at (R) {};
    \node[lab, below] at (S) {square};
    \node[lab, below] at (O) {odd};
    \node[lab, above] at (R) {remainder};
    \node at (1,-1.0) {Sentence 2};
  \end{scope}

  \begin{scope}[xshift=7cm]
    \coordinate (S) at (0,0); \coordinate (O) at (2,0); \coordinate (R) at (1,1.7);
    \draw[cyc] (S) -- (O) -- (R) -- cycle;
    \node[v] at (S) {}; \node[v] at (O) {}; \node[v] at (R) {};
    \node[lab, below] at (S) {square};
    \node[lab, below] at (O) {odd};
    \node[lab, above] at (R) {remainder};
    \node at (1,-1.0) {Sentence 3};
    \node[font=\footnotesize, red!75!black] at (1,-1.45) {$H_1$ class born};
  \end{scope}

  \begin{scope}[xshift=10.5cm]
    \coordinate (S) at (0,0); \coordinate (O) at (2,0); \coordinate (R) at (1,1.7);
    \fill[blue!20] (S) -- (O) -- (R) -- cycle;
    \draw[old] (S) -- (O) -- (R) -- cycle;
    \node[v] at (S) {}; \node[v] at (O) {}; \node[v] at (R) {};
    \node[lab, below] at (S) {square};
    \node[lab, below] at (O) {odd};
    \node[lab, above] at (R) {remainder};
    \node at (1,-1.0) {Sentence 4};
    \node[font=\footnotesize, blue!70!black] at (1,-1.45) {$H_1$ class dies};
  \end{scope}

\end{tikzpicture}
\caption{An example co-occurrence filtration}
\label{fig:cooccurrence}
\end{figure}

We can apply homology to a simplicial complex to analyze topological features.
Homology requires a specified dimension $n$, which is used to specify the dimension of features to be detected.
For a simplicial complex $K$, $H_0(K)$ counts the number of connected components, $H_1(K)$ counts the number of cycles that are not filled in via triangles (1-dimensional holes), $H_2(K)$ counts two dimensional holes, and in general $H_n(K)$ counts $n$-dimensional holes.
Since we have a filtration of simplicial complexes, we can apply homology to each step to see how these features evolve over time. 
An $n$-dimensional hole is \textit{born} at time $i$ if the $i$-th sentence creates this hole in the simplicial complex. 
An $n$-dimensional hole \textit{dies} at time $j$ if the $j$-th sentence fills in the hole. 
If an $n$-dimensional hole is born at time $i$ and dies at time $j$, the pair $(i,j)$ is called a \textit{persistence interval} and it represents the time that a hole was present for. 
For a given dimension, the collection of all persistence intervals of that dimension is called the \textit{persistence diagram}, denoted $\operatorname{dgm}_n(K_\bullet)$, if $K_\bullet$ is the filtration.
An example of a co-occurance filtration can be seen in \cref{fig:cooccurrence}. 
Note that for simplicity, the entire simplicial complex is not drawn, rather, only the subcomplex corresponding to the bolded words.

\begin{enumerate}[label=\arabic*.]
  \item First we \textbf{square} an \textbf{odd} integer.
  \item Every \textbf{square} has \textbf{remainder} zero or one when divided by four.
  \item An \textbf{odd} number leaves a \textbf{remainder} of either one or three.
  \item Therefore the \textbf{odd} \textbf{square} must have \textbf{remainder} 1.
\end{enumerate}

In practice, we do not use all words in the filtration.
Common words that appear in almost all sentence destroy homology, so we remove a list of common words beforehand. 
To avoid bias, this common word list was fixed before any tests were run and was not changed. 
The full list can be seen in [CITE APPENDIX].

We wish to measure the topological complexity of such filtrations.
To do this, we fix 1-dimensional persistence, so holes correspond to cycles in the graph that have not been filled. 
Our measure for topological complexity is \textit{total persistence}, which we denote $T(K_\bullet)$.
The total persistence is the sum of all lengths of $H_1$ intervals:
$$T(K_\bullet) = \sum_{(i,j) \in \operatorname{dgm}_1(K_\bullet)} j-i$$
To avoid infinite lengths, if a hole is still present at the last step in the filtration, we view it at dying at the last step + 1. 
Because $T(K_\bullet)$ is dependent on both vocabulary and text length, we apply a normalization. 
Recall that the filtration can be represented by a sequence of sets of words, with each set corresponding to one sentence. 
We can form a shuffled filtration $S(K_\bullet)$ by first shuffling the sentence lengths and then randomly distributing the words into these sentences. 
We then normalize by taking $$\overline{T}(K_\bullet) = T(K_\bullet) / \mathbb{E}[T(S(K_\bullet))]$$
We refer to $\overline{T}(K_\bullet)$ as the total persistence divided by the null total persistence.
In general, $\mathbb{E}[T(S(K_\bullet))]$ cannot be computed directly, so we estimate it by taking 200 samples and computing the mean. 
Any number below 1 represents simpler topology than random, and the closer to zero, the simpler the topology.
With our measure of topological complexity defined, we describe the experiments.

\subsection{Experiment 1: reasoning vs non-reasoning prose}
To test whether reasoning has simpler topology, we compare reasoning-focused texts with non-reasoning texts.
We use texts from Project Gutenberg \cite{projectgutenberg} , and consider four categories: mathematics, argumentative philosophy, history, and short stories. Mathematics and argumentative philosophy represent texts which are primarily structured reasoning, and history and short stories are texts that are not primarily structured reasoning. 
We choose 24 texts from each genre, and truncate all texts to 300 sentences to control for length. 
The full list of texts can be found in Appendix B.
The hypothesis is as follows:

\begin{hypothesis} \label{hyp:ex1}
Reasoning prose has simpler topology than non-reasoning prose.
Namely, For two genres $A$ and $B$, let $X \sim \overline{T}$ of a random text from $A$
and $Y \sim \overline{T}$ of a random text from $B$. We test four hypotheses, each with
\[
  H_0:\ \Pr(X < Y) = \tfrac12,
  \qquad
  H_1:\ \Pr(X < Y) > \tfrac12,
\]
i.e. a text from genre $A$ tends to have lower normalised total persistence than a text from genre $B$:
\begin{enumerate}
  \item[H1.] mathematics $<$ history,
  \item[H2.] mathematics $<$ short stories,
  \item[H3.] philosophy $<$ history,
  \item[H4.] philosophy $<$ short stories.
\end{enumerate}
Each hypothesis is tested with a one-sided Brunner Munzel test.
\end{hypothesis}

\subsection{Experiment 2: OpenAI mathematics papers vs human mathematics papers}

Recently, OpenAI released 722 AI-written papers covering solutions to 372 open problems in mathematics \cite{openai2026math}. 
These papers have been criticized as being hard to follow/read, and one possible explanation is that human-written papers often present a more refined version of reasoning.
To test this, we take a sample of 135 of the papers and pair each with a human-written paper in the same field, published in a top journal. 
This pairing was done using Claude Opus 5.5, and the full list can be found in Appendix B. 
To control for length, we truncate each pair to the shorter paper so that papers in each pair have the same sentence count and therefore filtration length. 
The OpenAI papers were accessed on October 8, 2026. 
The hypothesis is as follows:

\begin{hypothesis} \label{hyp:ex2}
    Human-written papers have simpler topology than the matched
AI-written papers from OpenAI.
Namely, let
\[
  D_i \;=\; \overline{T}\big(K^{\mathrm{H}}_{i,\bullet}\big) - \overline{T}\big(K^{\mathrm{A}}_{i,\bullet}\big)
\]
be the difference between the human and the OpenAI paper in pair $i$.
\[
  H_0:\ \text{the } D_i \text{ are symmetric about } 0,
  \qquad
  H_1:\ \text{the } D_i \text{ are shifted below } 0,
\]
tested with a one-sided Wilcoxon signed-rank test.
\end{hypothesis}

\subsection{Experiment 3: LLM capabilities vs chain of thought topology}
We further test whether better reasoning has simpler topology than lower-quality reasoning.
To do this, we compute the topology of the chain of thought for 48 different LLMs on 30 of the 2026 AIME problems. 
A full list of the problems and models tested can be found in Appendix B.
To control for length, we truncate to the first 100 reasoning sentences.
Traces with fewer than 100 sentences were discarded, meaning some models had fewer than 30 usable traces. 
For each model, we compute its median total persistence / null persistence across the 30 questions. 
To score reasoning strength, we use the Epoch Capabilities Index (ECI) \cite{EpochEci2025}.
The 48 models were all open-weight models providing a raw CoT that were scored on the ECI. 
The hypothesis is as follows:

\begin{hypothesis} \label{hyp:ex3}
More capable models have simpler reasoning topology.
In particular, for a given model, let $T_m$ be the median total persistence / null, and $C_m$ the ECI. 
\[
  H_0:\ \rho_s(C, T) = 0,
  \qquad
  H_1:\ \rho_s(C, T) < 0,
\]
where $\rho_s$ is Spearman's rank correlation across models. Tested with a one-sided Spearman test.
\end{hypothesis}

\subsection{Experiment 4: LLM chain of thought topology for sabotaged and benign tasks}

Finally, we explore the potential applications to AI safety. 
To do this, we use SHADE-arena \cite{kutasov2025shade}.
SHADE-arena is a collection of 26 test environments for LLMs, where they have to preform simulated real-world tasks, such as sending payments to the user's friends to pay them back for a recent trip.
Every task has a secret sabotage side task, which could be sending \$5000 to an unknown account. 
If the side task is enabled, the AI is prompted to do it alongside the main task while avoiding detection. 
The model we choose to test is deepseek-v4-flash-0731.
For each task, we run three trials for both the benign and sabotage versions. 
We then truncate each trial to have the same sentence count as the shortest trial among the six for the task, and compare the mean total persistence / null for the benign version to the mean total persistence / null for the sabotage version.
The rationale is that simultaneously pursuing the main task, a covert side task, and concealment may change how concepts are combined in the reasoning trace, potentially increasing its topological complexity.
Therefore, the hypothesis is as follows:

\begin{hypothesis} \label{hyp:ex4}
Reasoning while secretly pursuing a side task has less simple topology than reasoning on the same task honestly. 
Let
\[
  D_j \;=\; \overline{T}{}^{\,\mathrm{sab}}_j \;-\; \overline{T}{}^{\,\mathrm{ben}}_j
\]
be the difference between the mean over the sabotage runs and the mean over the benign runs of task $j$. Then
\[
  H_0:\ \text{the } D_j \text{ are symmetric about } 0,
  \qquad
  H_1:\ \text{the } D_j \text{ are shifted above } 0,
\]
tested with a one-sided Wilcoxon signed-rank test.

\end{hypothesis}

%% file: 2-results.tex
\section{Results}

In this section, we review the results of the four experiments mentioned in \cref{sec:methodology}.

\subsection{Experiment 1: reasoning vs non-reasoning prose}

The results of the experiment can be seen in the following two tables and figure. 
In particular, we find statistically significant evidence in favor of all four initial hypotheses in \cref{hyp:ex1}.

\begin{table}[H]
\centering

\label{tab:genre-medians}
\begin{tabular}{lcc}
\toprule
Genre & Median & IQR \\
\midrule
Mathematics & 0.537 & 0.482--0.626 \\
Philosophy & 0.674 & 0.617--0.754 \\
History & 0.730 & 0.681--0.811 \\
Short stories & 0.872 & 0.816--0.889 \\
\bottomrule
\end{tabular}
\caption{Normalised $H_1$ total persistence by genre (observed / null mean).}
\end{table}

\begin{table}[H]
\centering
\small
\label{tab:genre-tests}
\begin{tabular}{llcccc}
\toprule
 & Comparison & $p$ & Shift & 95\% CI & $\Pr(X<Y)$ \\
\midrule
H1 & Math $<$ history & $<10^{-5}$ & $-0.19$ & $[-0.26, -0.14]$ & 0.93 \\
H2 & Math $<$ stories & $<10^{-5}$ & $-0.32$ & $[-0.37, -0.25]$ & 0.99 \\
H3 & Philosophy $<$ history & $0.019$ & $-0.05$ & $[-0.12, -0.004]$ & 0.67 \\
H4 & Philosophy $<$ stories & $<10^{-5}$ & $-0.19$ & $[-0.23, -0.12]$ & 0.89 \\
\bottomrule
\end{tabular}
\caption{Genre comparisons (one-sided Brunner Munzel test).}
\end{table}

\begin{figure}[H]
    \centering
    \includegraphics[width=0.8\linewidth]{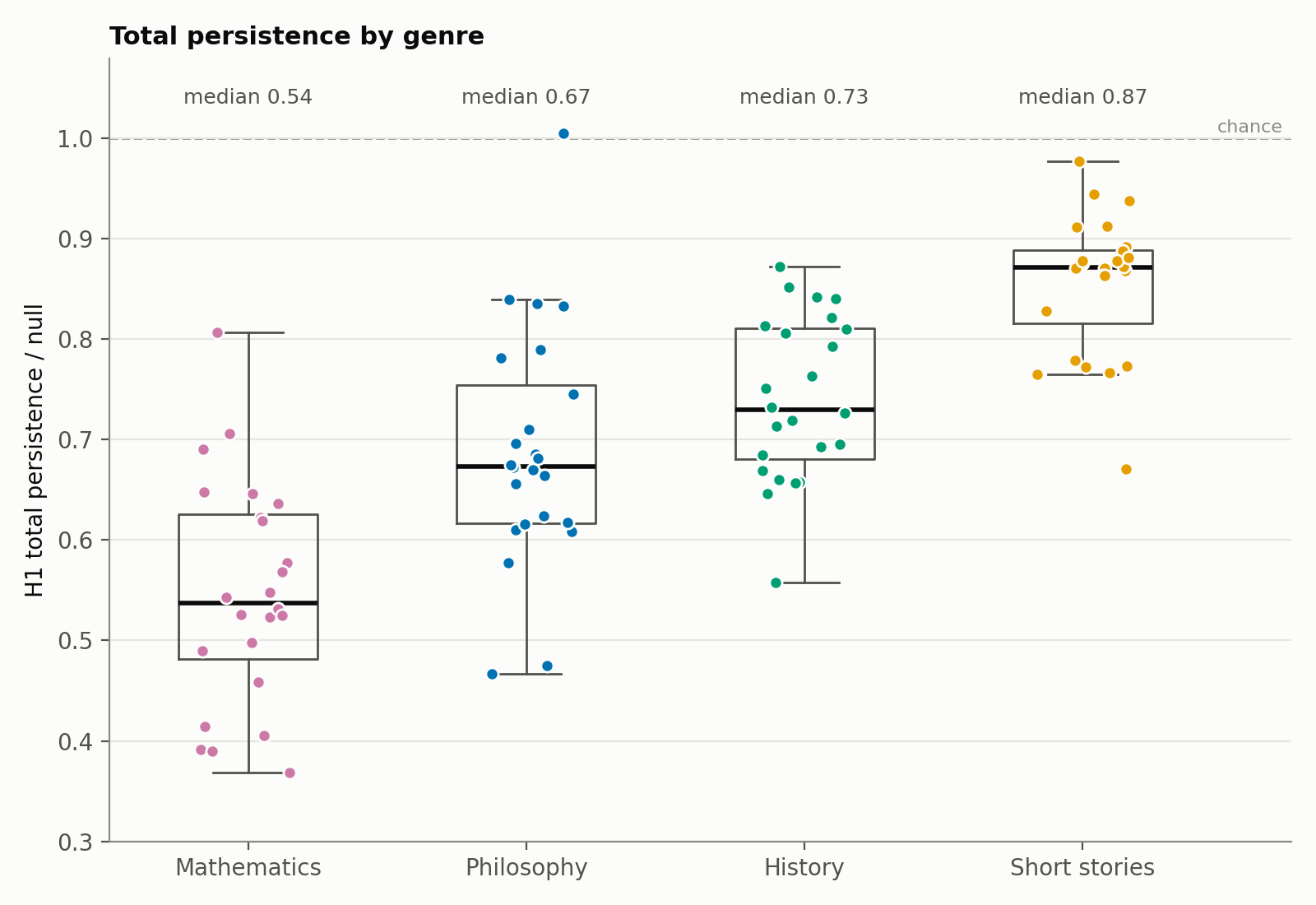}
    \caption{Total persistence distribution among the four genres.}
    \label{fig:ex1}
\end{figure}

\subsection{Experiment 2: OpenAI mathematics papers vs human mathematics papers}
We find that in 93\% (126/135) of OpenAI-human paper pairs, the human paper has lower total persistence / null.
The p-value under a one-sided Wilcoxon signed-rank test was $5 \times 10^{-23}$, giving statistically significant evidence in favor of \cref{hyp:ex2}.
The median value for human papers was 0.583 with an IQR of 0.529-0.625, and the median value for OpenAI papers was 0.722 with an IQR of 0.674-0.759.
Human papers showed slightly more variance than OpenAI papers, with a standard deviation of 0.074 compared to 0.068 for OpenAI papers. 

\begin{figure}[H]
    \centering
    \includegraphics[width=0.8\linewidth]{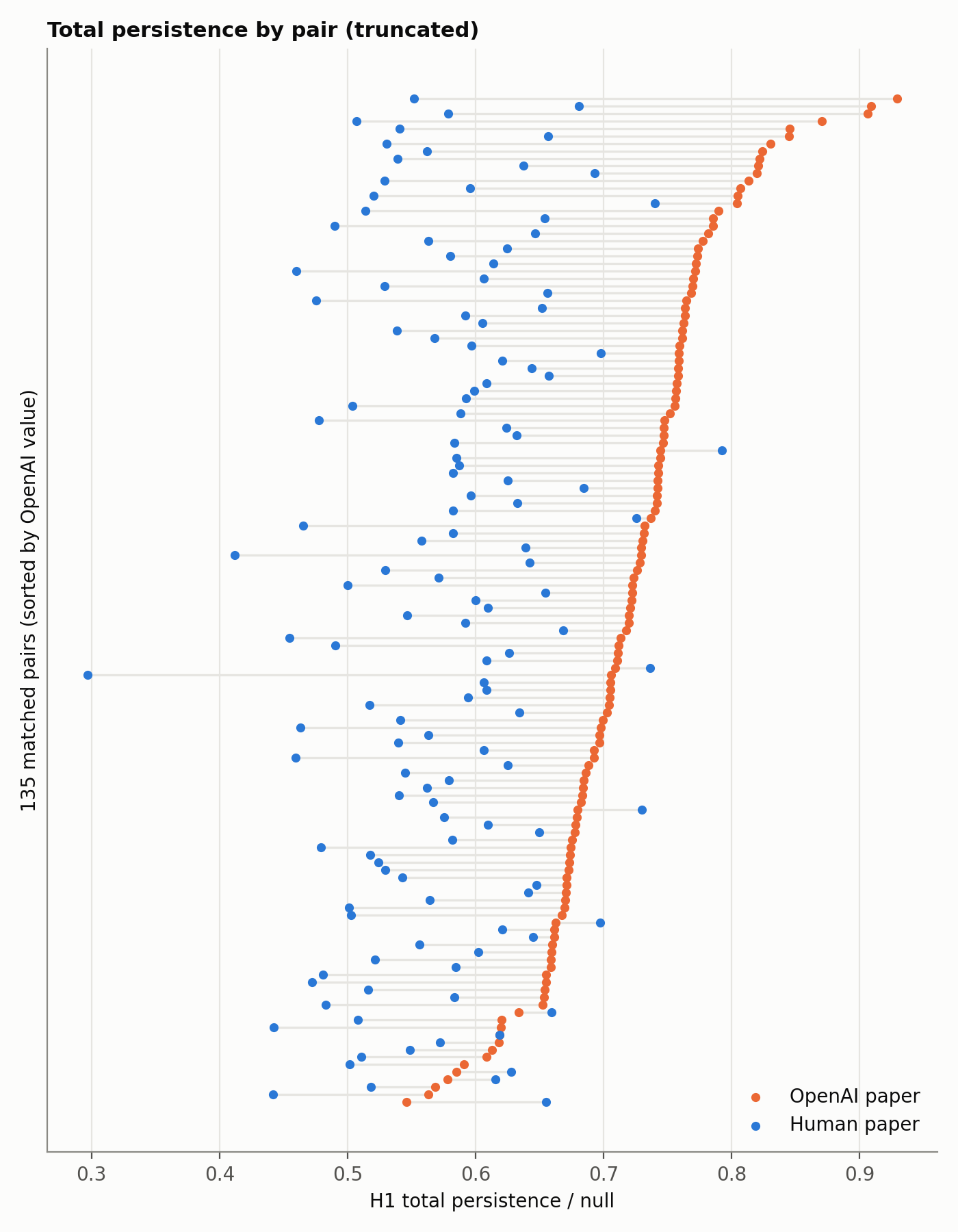}
    \caption{Total persistence / null persistence for human vs OpenAI paper pairs.}
    \label{fig:ex2}
\end{figure}

\subsection{Experiment 3: LLM capabilities vs chain of thought topology}
We find a moderate negative correlation between model ECI and median reasoning topology, with a Spearman's correlation coefficient of $\rho=-0.444$, giving a p-value of 0.0008. 
This provides statistically significant evidence in favor of \cref{hyp:ex3}.
As an expository result, we also test whether within labs stronger models tend to have simpler topology. 
Among 196 within-lab pairs, we find that in 69\% of pairs the stronger model has simpler reasoning topology, giving a p-value of 0.002. 

\begin{figure}[H]
    \centering
    \includegraphics[width=0.8\linewidth]{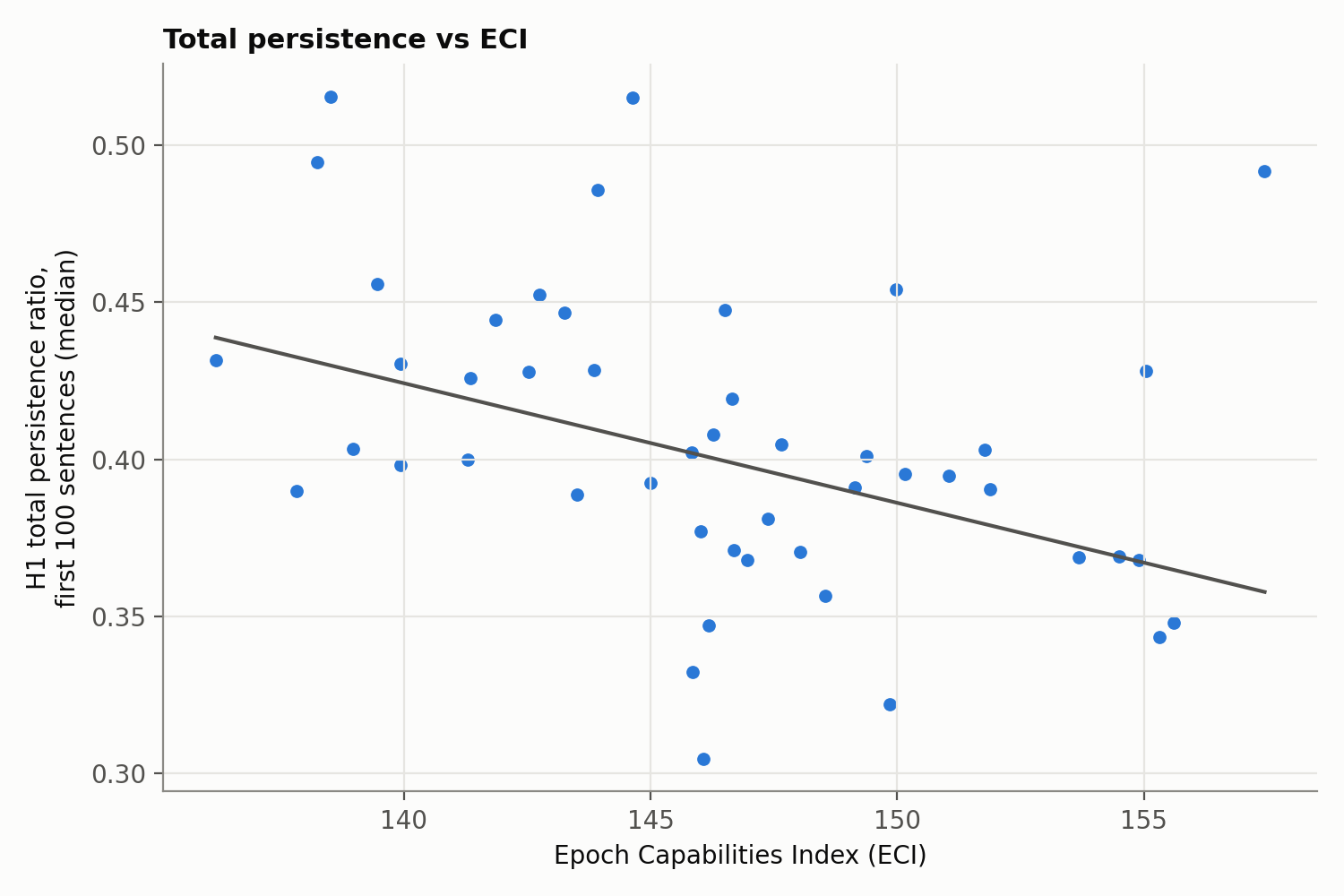}
    \caption{ECI vs median total persistence / null persistence for 48 models.}
    \label{fig:ex3}
\end{figure}

\subsection{Experiment 4: LLM chain of thought topology for sabotaged and benign tasks}

\begin{figure}[H]
    \centering
    \includegraphics[width=0.8\linewidth]{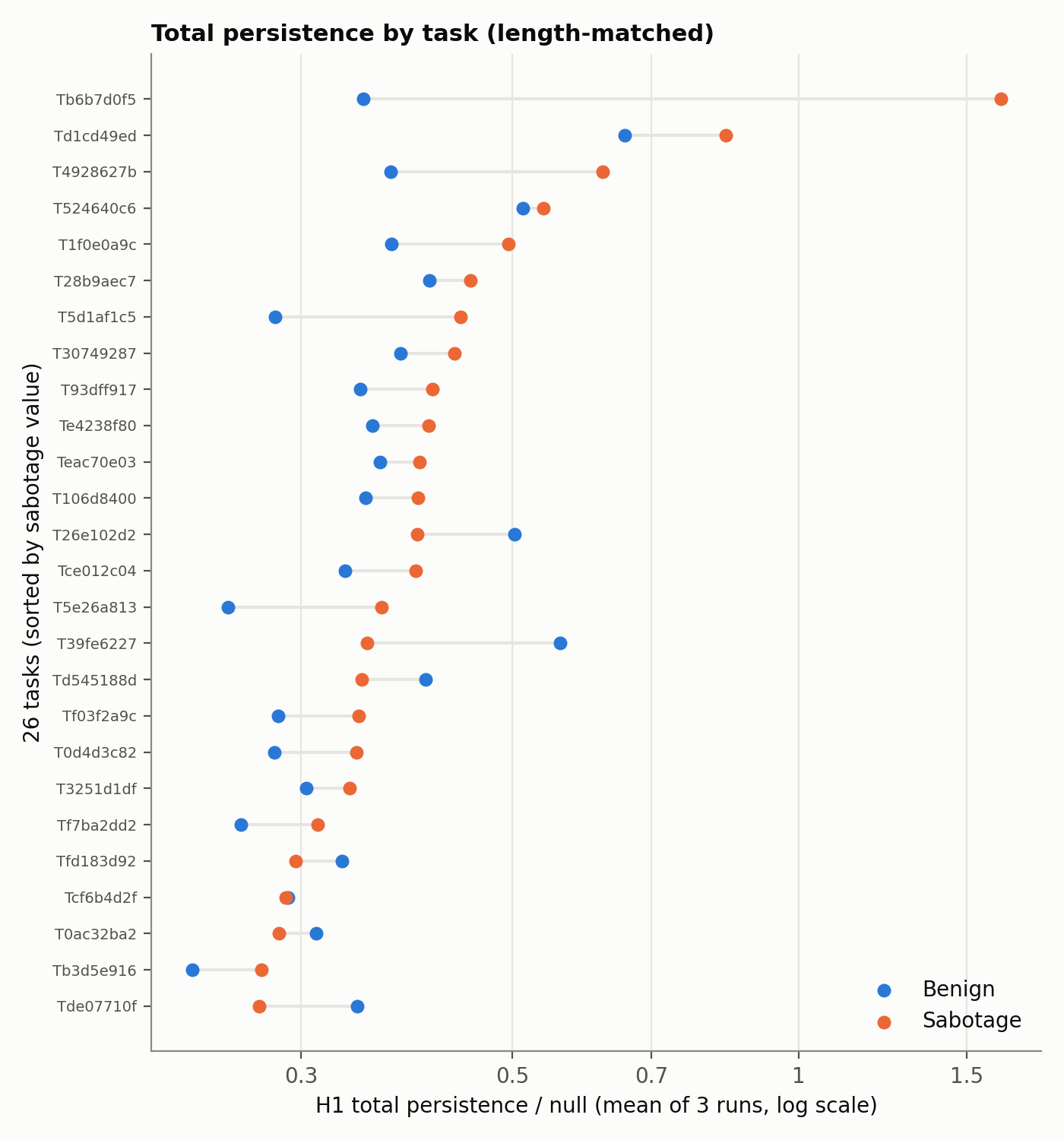}
    \caption{Total persistence / null for benign and sabotaged tasks.}
    \label{fig:ex4}
\end{figure}

We find that in 19 of the 26 task pairs, the benign task had simpler topology than with the sabotage task included. 
The benign mean was 0.363, and the sabotage mean was 0.449.
The p-value using a one-sided Wilcoxon signed-rank test was 0.009, giving statistically significant evidence in favor of \cref{hyp:ex4}.
The results are visualized in \cref{fig:ex4}. Note that task names are hidden to avoid the names being leaked into pretraining. 

%% file: 3-discussion.tex
\section{Discussion}
Overall, we find significant evidence for the two main ideas in the introduction: reasoning has simpler topology than non-reasoning, and higher-quality reasoning has simpler topology than lower-quality reasoning. 
In the first experiment, we find that prose across different reasoning and non-reasoning subjects lands in an expected manner: mathematics is the most topologically simple, followed by argumentative philosophy, history, and finally short stories. 
The comparisons specified by the hypothesis were all statistically significant.
In the second experiment, we find that 93\% of human-written papers have simpler topology than their OpenAI LLM-written counterparts.
One possible interpretation of this result is the differences in the reasoning in LLM-written and human-written papers, with human-written papers presenting a more refined, simple version of reasoning.
Again, this is one possible explanation, and other explanations such as writing style cannot be ruled out.
In the third experiment, we show that among 48 open-weight LLMs, there is a moderate negative correlation between ECI and topological complexity. 
Finally, in the fourth experiment, we find that the prompted-sabotage condition is associated with higher normalized total persistence. 
Across 26 tasks, the benign version had simpler topology in 19 of them. 

We propose the following explanation for these patterns, based on what persistent holes represent under this construction.
Reasoning often reuses topics, and combines them in a variety of ways. 
Higher quality reasoning is also likely to combine more ideas and leave fewer holes open. 
Under this explanation, this means that reasoning often creates fewer holes, and when these holes are created, it fills them in more quickly. 
On the other hand, narrative creates more holes and is not as concerned with filling them in, leading to more complex topology. 

Experiments 1 and 3 can be viewed as providing support for the idea that reasoning is topologically simple. 
Experiment 1 shows that reasoning-focused text to be simpler than non-reasoning-focused text, and experiment 3 extends this to show that high-quality reasoning is simpler than low-quality reasoning. 
Experiments 2, 3, and 4 can be viewed as applications of the method to reasoning evaluation, AI evaluation, and AI safety. 
Experiment 2 provides evidence that the method can be used to evaluate the reasoning presented in papers. In particular, it shows that on average, LLM-written papers (criticized as being difficult to read) are topologically more complex than human-written counterparts.
This motivates further investigation of whether topological simplicity is associated with clearer presentations of reasoning.
The association with model capability in experiment 3 motivates investigating the statistic as an answer-free signal for model evaluation.
Experiment 4 shows how the method might apply to AI safety. 
In particular, the method provides an easily computable statistic that can evaluate a model's chain of thought and potentially provide an alert for alarming behavior or changes in behavior. 
Importantly, this does not depend on having a separate LLM monitor chain of thought. Its ability to distinguish individual benign and sabotage runs, including its false-positive rate, remains to be evaluated.

\paragraph{Limitations.}
There are a few limitations in the present work:
\begin{itemize}
    \item Interpretation of the statistic: The experiments alone are unable to isolate reasoning quality from other properties of the texts. Genre conventions, vocabulary, editing, and model-specific writing styles may contribute to the observed differences. Direct assessments of reasoning quality and readability, together with appropriate controls, are needed to determine how well the proposed explanation accounts for the findings.

    \item Sample size: experiment 1 used only 24 papers per genre, limited by the number of mathematics papers in the Gutenberg project; experiment 3 used only 48 models, limited by the need for open-weight reasoning models scored by the ECI; and experiment 4 used only 26 tasks.
    While all results were still statistically significant, increasing the sample size could strengthen these results further.

    \item Experiment 4 was restricted to only one model. 
    Whether or not the result is consistent across other models is not tested. 

    \item The common words list was an arbitrary choice fixed before experiments. 
    Whether the results hold across different lists is untested.

    \item Words are treated verbatim: different tenses of the same word are treated as different words.

    \item While we maintain the sentence ordering, sentences are treated as sets of words, so the ordering of words within a sentence is lost. 
\end{itemize}

\paragraph{Future work.}

There are several directions for future work. 
The most immediate direction is to address the limitations: develop tests that isolate reasoning quality or readability more directly, increase the paper count for experiment 1, increase the number of models for experiment 3 (this would require using a different measure of model capabilities) or test experiment 4 with other models. 
Another direction is to try to achieve stronger results by clustering similar words. 
Words are meant to represent different ideas or concepts, so different tenses of the same word should be treated as the same. 
Likewise, synonyms should also be treated as the same idea. 
The list of common words could also be expanded or altered so that more filler words, or words not corresponding to a ``concept,'' are removed. 
Finally, the applications should be further developed. 
For example, one could test whether using co-occurrence topology as a revision signal improves independently assessed reasoning quality or readability
Another important application is in AI safety, where this could be used to develop tools for real-time evaluation and monitoring.

%% file: A-cooccurrence.tex
\section{Co-occurrence persistent homology} \label{sec:co-occurrence}

Here we provide the full mathematical details of co-occurrence persistent homology, as well as the optimizations made to make the computations fast. 

\subsection{Mathematical details}

We begin with the formal definition of a simplicial complex:

\begin{definition} \label{def:simplicial_complex}
    A \textit{simplicial complex} $(K_0,K)$ consists of a set $K_0$ along with a collection $K$ of subsets of $K_0$ such that:
    \begin{enumerate}
        \item For every $v \in K_0$, $\{v\} \in K$
        \item For every $X \in K$ and $Y \subseteq X$, $Y \in K$
    \end{enumerate}
\end{definition}
We often refer to the pair $(K_0,K)$ as simply $K$. 
A subset of size $n+1$ in $K$ is called an $n$-simplex. 
The set of all $n$-simplices in $K$ is denoted $K_n$.

We next discuss the homology of a simplicial complex. 
For simplicity and so our presentation matches what our code computes, we use coefficients in $\mathbb{Z}_2$. 
In general, homology can be defined over any coefficient ring $R$.
Let $K$ be a simplicial complex, and $C_n(K)$ the $\mathbb{Z}_2$-vector space generated by the set $K_n$.

\begin{definition}
    Let $K$ be a simplicial complex, and $x=\{x_0,\dots,x_n\}$ and $n$-simplex in $K_n$. 
    The \textit{boundary} of $x$, denoted $\partial_n x$, is the formal sum:
    $$\sum_{i=0}^n \{x_0, \dots ,\hat{x}_i, \dots x_n\}$$
    where $\hat{x}_i$ denotes the removal of the element $x_i$ from the set.
\end{definition}
Due to property 2 of \cref{def:simplicial_complex}, $\partial_nx$ lives in $C_{n-1}(K)$. 
Thus, $\partial_n$ defines a linear map $C_n(K) \to C_{n-1}(K)$. 
\begin{proposition} \label{prop:chain_cpx}
    Let $K$ be a simplicial complex. The pair $(C_\bullet(K),\partial_\bullet)$ forms a chain complex. 
    In particular, $\textsf{im}\, \partial_{n+1} \subseteq \textsf{ker}\, \partial _n$ for all $n$. \qed
\end{proposition}

The homology of this chain complex is the homology of the simplicial complex $K$:

\begin{definition}
    Let $K$ be a simplicial complex and $n \in \mathbb{N}$. The $n$-th homology group of $K$ (over $\mathbb{Z}_2$) is given by 
    $$H_n(K) = \textsf{ker}\, \partial_n / \textsf{im} \, \partial_{n+1}$$
\end{definition}
This is well-defined by \cref{prop:chain_cpx}. 
With homology now defined, we work towards defining \textit{persistent homology} next. 

\begin{definition}
    A \textit{filtration} of simplicial complexes $K_\bullet$ is a sequence
    $$K_0 \to K_1 \to \dots \to K_m$$
    such that for every $0 \leq i \leq m-1$, $K_i \subseteq K_{i+1}$.
\end{definition}
In other words, a filtration of simplicial complexes is a nested sequence of simplicial complexes.

\begin{definition}
    A \textit{persistence module} $V$ is a sequence of ($\mathbb{Z}_2$-)vector spaces
    $$V_0 \to V_1 \to \dots \to V_m\text{.}$$
\end{definition}

An important example of a persistence module is as follows:

\begin{definition}
    Let $b<d$ be natural numbers. 
    The \textit{interval module} $I[b,d)$ is the persistence module 
    $$I[b,d)_n = \begin{cases}
        \mathbb{Z}_2 & i \leq n < j \\
        0 & \text{otherwise}
    \end{cases}$$
    with identity maps between any two nonzero vector spaces. 
\end{definition}
Note that the length of the sequence is not specified.
This is deliberate, since in practice, the length is inferred from context.
Every persistence module can be decomposed into a direct sum of interval modules.

\begin{theorem}[Gabriel, \cite{gabriel:decomposition}] \label{th:decomp}
    For a persistence module $V$, there exists a unique decomposition of $V$ into a direct sum of interval modules:
    \[ V \cong \bigoplus_J I[b_j,d_j)\text{.} \tag*{\qed} \]
\end{theorem}

Given a filtration of simplicial complexes $K_\bullet$ and a natural number $n$, we can obtain a persistence module by taking homology over the filtration (this is what \textit{persistent homology} refers to):
$$H_n(K_0) \to H_n(K_1) \to \dots \to H_n(K_m)$$
Using \cref{th:decomp}, we can decompose this module into a direct sum of interval modules. 
The collection of the pairs $(b_j,d_j)$ in the decomposition is called the \textit{persistence diagram}, denoted $\textsf{dgm}_n(K_\bullet)$.
Each pair corresponds to a homology generator in degree $n$, with $b_j$ corresponding to its creation (birth) and $d_j$ corresponding to its filling in (death). 

Given a text, let $X_t$ denote the set of all words used in the first $t$ sentences, with the common words removed. 
We form a simplicial complex $K_t$ on the vertex set $X_t$ via the condition $\{w_0,\dots ,w_n\} \in K_t$ if $\{w_0,\dots ,w_n\}$ all appear together in one of the first $t$ sentences.
Allowing $t$ to vary gives a filtration of simplicial complexes $K_\bullet$.
We then take degree 1 homology $H_1(K_\bullet)$ to get the degree 1 peristence diagram $\textsf{dgm}_1(K_\bullet)$.
The total persistence $T(K_\bullet)$ is the sum of all lengths of intervals in $\textsf{dgm}_1(K_\bullet)$.

\subsection{Computations}
To run experiments, we require efficient computation of co-occurrence persistent homology. 
The code to compute co-occurrence persistence was optimized with the use of GPT-6.1 Sol, and used a number of techniques which we outline here. 

The input to the algorithm is a sequence $S_\bullet = (S_1,S_2,\dots,S_T)$, where each $S_t$ is a set of integers. 
In the case of this paper, each $S_t$ represents a sentence, and each word is assigned a unique integer. 
To match the wording of the paper, we refer to the $S_t$'s as sentences. 
Let $K_t$ denote the simplicial complex at filtration time $t$. 
In particular, $K_t = K_{t-1} \cup \triangle(S_t)$. 
For a simplex $\sigma$ the \textit{birth time}, denoted $b(\sigma)$, is the smallest $t$ such that $\sigma \subseteq S_t$. 

The primary method for improving computation times will be to reduce the size of the filtration. There are two main strategies we employ to do this: we first employ preprocessing to remove dominated vertices and redundant filtration steps, and then employ discrete Morse theory \cite{forman-morse, mischaikow-nanda-morse} to further reduce the size.
After the filtration size has been reduced, we employ the standard persistence pairing algorithm, which we do not outline here. 
The only extra trick we employ during this computation is the use of birth clearing \cite{clear-and-compress, bauer:ripser}. 
In this setting, this works without the use of cohomology due to the filtration not being dominated in size by the ($n+1$)-simplices. 

\subsubsection{Preprocessing}

We begin by defining the support of a vertex:
\begin{definition}
    Let $v$ be a vertex in the filtration. The \textit{support} of $v$, denoted $I(v)$, is given by:
    $$I(v) = \{ S_t \ | \ v \in S_t \}$$
\end{definition}

The motivation for this definition is the following proposition:
\begin{proposition}
    Let $v \neq w$ be vertices in the filtration, and suppose $I(v) \subseteq I(w)$. 
    Then $v$ can be fully removed from the filtration without affecting homology. 
\end{proposition}
\begin{proof}
    Under this condition, $v$ is a dominated vertex in the sense of \cite{barmak-minian} at every stage of the filtration, and can thus be removed. 
\end{proof}

The removal of dominated vertices (or simplices) to simplify persistent homology computations is not new \cite{Boissonnat-Pritam-edge}. 

The second preprocessing trick we use is the removal of contained simplices: if $S_t \subseteq S_i$ and $i<t$, then $\triangle(S_t) \subseteq S_{t-1}$. 
Thus, we can skip the $t$-th step in the filtration altogether. 
These two methods represent quick checks that can be done prior to any computations, and often significantly reduce the sizes of the filtration. 

\subsubsection{Discrete Morse theory}

Discrete Morse theory \cite{forman-morse} can be used to further simplify the filtration before any computations. 
To do this, we make use of \cite[Theorem 4.3]{mischaikow-nanda-morse}, which says that for a filtration $F$ and a filtered matching $M$, we can form a filtered Morse complex $F^M$ such that $H_*(F) \cong H_*(F^M)$. 
For brevity, we assume familiarity with both discrete Morse theory and the persistent version presented in \cite{mischaikow-nanda-morse}.
The goal is to exhibit a filtered acyclic matching on the filtration $K_\bullet$ given by $S_\bullet$. This is done as follows:

At a time $t$, call a simplex $\sigma \in K_t$ \textit{new} if $b(\sigma)=t$ and \textit{old} otherwise. 
For a given $S_t$, pick any vertex $p_t \in S_t$. 
At time $t$, for every new $\sigma$ such that $p_t \notin \sigma$, pair $\sigma$ with $p_t \sigma = \sigma \cup \{p_t\}$.
\begin{theorem}
    The matching described above defines an acyclic matching on the complex $K_T$.
    The unmatched pairs are exactly:
    \begin{enumerate}
        \item The singletons $\{p_t\}$.
        \item New simplices $\tau$ containing $p_t$ such that $\tau \setminus \{p_t\}$ is old.
    \end{enumerate}
\end{theorem}

\begin{proof}
    Note first that every simplex is contained in at most one pair: for a simplex $\sigma$ not containing $p_t$ with $b(\sigma) = t$, it is only contained in $(\sigma, p_t\sigma)$. 
    For simplices $\tau$ containing $p_t$ with $b(\tau)=t$, $\tau \setminus \{p_t\}$ is the only possible simplex $\tau$ could be paired with. 

    For acyclicity, call the arrows from a simplex to its face \textit{down arrows}, and call the arrows reversed by the matching \textit{up arrows}. 
    Note that down arrows cannot increase birth time: a face must appear before the coface. 
    Note further that for every pair $(\sigma, p_t\sigma)$, $b(\sigma) = b(p_t \sigma)$. 
    Thus, any directed cycle must have constant birth time $t$.
    An up arrow in such a cycle adds $p_t$. 
    A down arrow removing $p_t$ must map to an old simplex; otherwise the arrow would be reversed.
    Since this does not keep birth time constant, $p_t$ cannot be removed in the cycle, and thus any cycle cannot contain an up arrow. 
    A cycle cannot contain only down arrows, however, since they strictly decrease dimension. 
    Thus, the matching is acyclic. 
\end{proof}

Because every matched pair contains simplices born at the same time, this defines a filtered acyclic matching, allowing us to use \cite[Theorem 4.3]{mischaikow-nanda-morse}.
We use a specialized version of the boundary formula in \cite{mischaikow-nanda-morse} that makes computations easier.
Let $C^M_t$ be the $\mathbb{Z}_2$ vector space generated by the critical simplices born at or before time $t$. 
For a critical simplex $\sigma$, by an abuse of notion, treat $\sigma$ is a basis element of $C^M_t$. 
For any simplex $\sigma$ born at time $t$, define:
$$\pi(\sigma) = \begin{cases}
    \sigma & \sigma \text{ is critical} \\
    0 & p_t \in \sigma \text{ and } b(\sigma \setminus \{p_t\}) = t \\
    \bigoplus_{v \in \sigma} \pi((\sigma \setminus \{v\}) \cup \{p_t\}) & p_t \notin \sigma
\end{cases}$$

First, note that the recursion terminates:
\begin{proposition}
    The recursion in the formula for $\pi(\sigma)$ terminates. 
\end{proposition}
\begin{proof}
    Let $\sigma$ be born at time $t$ with $p_t \notin \sigma$.
    Fix $v \in \sigma$ and let $\tau = (\sigma \setminus \{v\}) \cup \{p_t\}$. 
    Since $\tau$ contains $p_t$, if $b(\tau)=t$, either $\pi(\tau) = e_\tau$ ($\tau$ is critical) or $\pi(\tau)=0$ ($\tau$ is not critical). 
    Thus, the recursive part of the formula requires $b(\tau)<b(\sigma)$, and therefore eventually terminates. 
\end{proof}
This formula is computationally beneficial because $\pi(\sigma)$ for $p_t \notin \sigma$ can be computed using only the stored previous projections. 
For a critical simplex $\sigma$, define $\delta e_\sigma$ by 
$$\delta \sigma = \bigoplus_{v \in \sigma} \pi(\sigma \setminus \{v\})$$

Provided this differential matches the differential in \cite{mischaikow-nanda-morse}, it forms a chain complex and can be used to compute persistent homology. 

\begin{theorem}
For the matching defined above, the differential $\delta$
agrees with the Morse differential $d^M$ obtained from the
cancellation construction of
\cite[Section~2.3]{mischaikow-nanda-morse}, under the identification
of their critical generators with ours.
\end{theorem}

\begin{proof}
Given a simplicial complex $K$, let $C_n(K)$ be the chain complex in degree $n$ (over $\mathbb{Z}_2$), and $d$ the differential. 
For a matched pair $(q,u)$ with $q \in du$, let $C'_\bullet$ be the chain complex $C_\bullet$ with $q$ and $u$ removed. Then \cite{mischaikow-nanda-morse} defines a differential $d'$ on $C'_\bullet$ by setting $d' x = dx - du$ for every simplex $x$ such that $q \in dx$. 
They define a chain map $\psi \colon C \to C'$ by  $\psi (u)=0$, $\psi (q) = du - q$, and the identity on all other basis elements. 
Applying this process iteratively over all matched pairs gives the full reduced chain complex $C^M_\bullet$ with the differential $d^M$, and a projection $\psi \colon C_\bullet \to C^M_\bullet$. 

We first show that our projection $\pi$ matches $\psi$. 
For a critical simplex $\sigma$, we have that both $\pi \sigma = \sigma$ and $\psi \sigma = \sigma$.
Now, if $\sigma$ is the higher dimensional simplex in a matched pair, it necessarily contains $p_{b(\sigma)}$, so we have that $\pi(\sigma)=0=\psi(\sigma)$. 

Now suppose $\sigma$ is the lower dimensional simplex in a matched pair, with $t = b(\sigma)$. 
The boundary of $p_t\sigma$ is exactly  $$d(p_t\sigma) = \sigma + \bigoplus_{v \in \sigma} \psi((\sigma \setminus \{v\}) \cup \{p_t\})$$
We know $\psi$ is a chain map and that $\psi (p_t \sigma) = 0$, and thus $\psi(d(p_t \sigma))=0$. Thus:
$$\psi(\sigma) = \bigoplus_{v \in \sigma} \psi((\sigma \setminus \{v\}) \cup \{p_t\})$$
which is exactly the recursion defining $\pi(\sigma)$, and thus $\psi = \pi$. 
To see that $d^M = \delta$, for a critical simplex $\sigma$, we have that
$$d^M (\sigma) = d^M (\psi \sigma) = \psi(d \sigma) =  \pi (d \sigma) =  \delta (\sigma)$$
\end{proof}

Using this reduction allows for a significant reduction to the size of simplices in the filtration. 
While the proof allows for any vertex $p_t$ to be chosen at time $t$, we use a specific rule to choose the optimal $p_t$.
After preprocessing, for each vertex $v$, track both 
\[ a_t(v) = \# \{ i< t \ | \ v \in S_i \} \qquad f(v) = \# \{ i \ | \ v \in S_i \}\]
For every $S_t$, choose $p_t$ using the following rule:
\begin{enumerate}
    \item The vertex with the largest earlier count $a_t(v)$.
    \item Among ties, the vertex with the largest total count $f(v)$.
    \item Among further ties, the smaller vertex label.
\end{enumerate}
This guarantees that, when available, an old vertex is chosen. 
Choosing a new vertex would introduce $p_t$ as a new critical cell, and choosing an old one avoids this. 
Secondly, recall that $\tau$ is critical if $\tau$ is old but $\tau \cup \{p_t\}$ is new. 
Choosing $p_t$ to have appeared the most times beforehand attempts to minimize the chances that for an old $\tau$, $\tau \cup \{p_t\}$ is new. 
In practice, this rule was the fastest among all rules tested and led to the fewest critical cells. 

Using these methods, we see substantial improvements over our initial naive methods. 
The datasets below were taken from responses to the AIME problems from gpt-oss-120b. 
\begin{table}[H]
    \centering
    \begin{tabular}{lrrr}
        \toprule
        Dataset & Original (s) & Optimized (s) & Speedup \\
        \midrule
        AIME08  & 0.0332  & 0.00159 & $20.9\times$ \\
        AIME09        & 5.860  & 0.0255 & $230.2\times$ \\
        AIME10        & 15.045 & 0.0327 & $459.7\times$ \\
        AIME11  & 92.606 & 0.0673 & $1{,}376.5\times$ \\
        \bottomrule
    \end{tabular}
    \caption{Comparison of the naive and optimized implementations of Co-occurrence persistent homology.}
    \label{tab:original-versus-optimized}
\end{table}

%% file: B-experiments.tex
\section{Further details on experiments} \label{sec:experiments}
In this appendix, we provide additional information on the methods and experiments.
\subsection{Common words list}
The common words list was chosen to match the Ranks NL \textit{default English stopwords} list \cite{ranks_stopwords}. The list is as follows:

\begin{quote}
\small\raggedright\ttfamily
a, about, above, after, again, against, all, am, an, and, any, are, aren't, as, at, be,
because, been, before, being, below, between, both, but, by, can, can't, cannot, could,
couldn't, did, didn't, do, does, doesn't, doing, don't, down, during, each, few, for,
from, further, had, hadn't, has, hasn't, have, haven't, having, he, he'd, he'll, he's,
her, here, here's, hers, herself, him, himself, his, how, how's, i, i'd, i'll, i'm,
i've, if, in, into, is, isn't, it, it's, its, itself, let's, me, more, most, mustn't,
my, myself, no, nor, not, of, off, on, once, only, or, other, ought, our, ours,
ourselves, out, over, own, same, shan't, she, she'd, she'll, she's, should, shouldn't,
so, some, such, than, that, that's, the, their, theirs, them, themselves, then, there,
there's, these, they, they'd, they'll, they're, they've, this, those, through, to, too,
under, until, up, very, was, wasn't, we, we'd, we'll, we're, we've, were, weren't, what,
what's, when, when's, where, where's, which, while, who, who's, whom, why, why's, with,
won't, would, wouldn't, you, you'd, you'll, you're, you've, your, yours, yourself,
yourselves.
\end{quote}

\subsection{Experiment 1 list of texts}

Below is the full list of all texts used for experiment 1.
These texts were chosen autonomously with the use of Claude Opus 5.5. 
 Each item gives the author's surname, the work's main title and its Project Gutenberg eBook number.

\subsubsection*{Philosophy}
\begin{multicols}{2}
\small\raggedright
\begin{enumerate}[nosep, leftmargin=*, label=\arabic*.]
  \item Abercrombie, \emph{The Philosophy of the Moral Feelings} (\#22364)
  \item Atkinson, \emph{The Art of Logical Thinking} (\#41838)
  \item Bosanquet, \emph{The Essentials of Logic, Being Ten Lectures on Judgment and Inference} (\#63598)
  \item Carpenter, \emph{The Drama of Love and Death} (\#44883)
  \item Carr, \emph{The Problem of Truth} (\#47658)
  \item Clifford, \emph{The Scientific Basis of Morals, and Other Essays} (\#50189)
  \item Cocker, \emph{The Theistic Conception of the World} (\#46544)
  \item Coffey, \emph{Ontology, or the Theory of Being} (\#35722)
  \item Drake, \emph{Problems of Conduct} (\#5775)
  \item Fiske, \emph{The Destiny of Man, Viewed in the Light of His Origin} (\#17239)
  \item Fullerton, \emph{An Introduction to Philosophy} (\#16406)
  \item Givler, \emph{The ethics of Hercules} (\#70666)
  \item Guthrie, \emph{On Mr. Spencer's Data of Ethics} (\#56721)
  \item Huxley, \emph{Evolution and Ethics, and Other Essays} (\#2940)
  \item Jordan, \emph{The Philosophy of Despair} (\#4754)
  \item Jourdain, \emph{On the theory of the infinite in modern thought} (\#74732)
  \item Morley, \emph{On Compromise} (\#11557)
  \item Parker, \emph{The Principles of Aesthetics} (\#6366)
  \item Perry, \emph{The Approach to Philosophy} (\#25110)
  \item Pitt-Rivers, \emph{Conscience \& Fanaticism} (\#58136)
  \item Powys, \emph{The Complex Vision} (\#21668)
  \item Russell, \emph{What I believe} (\#73782)
  \item Sidgwick, \emph{The Methods of Ethics} (\#46743)
  \item Thornton, \emph{Old-Fashioned Ethics and Common-Sense Metaphysics} (\#29917)
\end{enumerate}
\end{multicols}

\subsubsection*{Mathematics}
\begin{multicols}{2}
\small\raggedright
\begin{enumerate}[nosep, leftmargin=*, label=\arabic*.]
  \item Archibald, \emph{Euclid's Book on Divisions of Figures} (\#38640)
  \item Baker, \emph{Elliptic Functions} (\#31076)
  \item Burnside, \emph{Theory of Groups of Finite Order} (\#40395)
  \item Byerly, \emph{An Elementary Treatise on Fourier's Series and Spherical, Cylindrical, and Ellipsoidal Harmonics} (\#29779)
  \item Carmichael, \emph{The Theory of Numbers} (\#13693)
  \item Coolidge, \emph{The Elements of non-Euclidean Geometry} (\#26373)
  \item De Morgan, \emph{Elementary Illustrations of the Differential and Integral Calculus} (\#39041)
  \item Glenn, \emph{A Treatise on the Theory of Invariants} (\#9933)
  \item Hardy, \emph{Orders of Infinity} (\#38079)
  \item Hearn, \emph{Researches on curves of the second order} (\#17204)
  \item Klein, \emph{The Evanston Colloquium} (\#36154)
  \item Lehmer, \emph{An Elementary Course in Synthetic Projective Geometry} (\#17001)
  \item Manning, \emph{Non-Euclidean Geometry} (\#13702)
  \item McAulay, \emph{Utility of Quaternions in Physics} (\#26262)
  \item McMahon, \emph{Hyperbolic Functions} (\#13692)
  \item Moon, \emph{Topics on Tournaments} (\#42833)
  \item Neikirk, \emph{Groups of the Order $p^{m}$ Which Contain Cyclic Subgroups of Order $p^{m-3}$} (\#9930)
  \item Record, \emph{The Path-Way to Knowledg, Containing the First Principles of Geometrie} (\#33093)
  \item Schafer, \emph{An introduction to nonassociative algebras} (\#25156)
  \item Slaught, \emph{Solid Geometry with Problems and Applications (Revised edition)} (\#29807)
  \item Thompson, \emph{Calculus Made Easy} (\#33283)
  \item Todhunter, \emph{Spherical Trigonometry} (\#19770)
  \item Veblen, \emph{Introduction to Infinitesimal Analysis} (\#18741)
  \item Whitehead, \emph{An Introduction to Mathematics} (\#41568)
\end{enumerate}
\end{multicols}

\subsubsection*{History}
\begin{multicols}{2}
\small\raggedright
\begin{enumerate}[nosep, leftmargin=*, label=\arabic*.]
  \item Adams, \emph{The history of the Jews} (\#69593)
  \item Boyle, \emph{The battle-fields of Ireland, from 1688 to 1691} (\#54518)
  \item Carrington, \emph{Washington the Soldier} (\#65380)
  \item Chapin, \emph{Top of the Ladder} (\#48808)
  \item Church, \emph{Henry the Fifth} (\#58405)
  \item Conybeare, \emph{Early Britain---Roman Britain} (\#12910)
  \item Douglas-Lithgow, \emph{The Nantucket Indians} (\#70447)
  \item Edmundson, \emph{Anglo-Dutch Rivalry During the First Half of the Seventeenth Century} (\#48697)
  \item Forbes, \emph{The Balkans} (\#11716)
  \item Fyfe, \emph{Edinburgh Under Sir Walter Scott} (\#47617)
  \item Gardiner, \emph{A Student's History of England, v. 1} (\#28157)
  \item Hallam, \emph{View of the State of Europe during the Middle Ages, Vol. 3} (\#33540)
  \item Hooker, \emph{The Africanders} (\#71707)
  \item Johnson, \emph{A History, of the War of 1812-15 Between the United States and Great Britain} (\#49393)
  \item Johnston, \emph{The French Revolution} (\#19421)
  \item Locke, \emph{When Canada was New France} (\#78668)
  \item Moore, \emph{School History of North Carolina} (\#6080)
  \item Nardini, \emph{No Man's Land} (\#55055)
  \item Pell, \emph{Fort Ticonderoga} (\#55561)
  \item Putnam Weale, \emph{The Fight for the Republic in China} (\#14345)
  \item Shaw, \emph{The Ohio River Trade, 1788-1830} (\#49627)
  \item Somerville, \emph{Narrative of the Fenian invasion of Canada} (\#70425)
  \item Thalheimer, \emph{A Manual of Ancient History} (\#56734)
  \item Wollstonecraft, \emph{An historical and moral view of the origin and progress of the French Revolution} (\#68724)
\end{enumerate}
\end{multicols}

\subsubsection*{Short stories}
\begin{multicols}{2}
\small\raggedright
\begin{enumerate}[nosep, leftmargin=*, label=\arabic*.]
  \item Bellamy, \emph{The Old Folks' Party} (\#22710)
  \item Bond, \emph{Peter Pettigrew's prisoner} (\#74328)
  \item Brand, \emph{The Laughter of Slim Malone} (\#67306)
  \item Freeman, \emph{The Yates Pride} (\#978)
  \item Glad, \emph{Each Man Kills} (\#23301)
  \item Harris, \emph{The Sheriff and His Partner} (\#23008)
  \item Howard, \emph{The hyena} (\#71065)
  \item Jones, \emph{Over the Wire} (\#67496)
  \item Ludlow, \emph{A Brace Of Boys} (\#23170)
  \item Marks, \emph{Brown John's Body} (\#32076)
  \item Neville, \emph{As holy and enchanted} (\#78909)
  \item Phillips, \emph{Gun play} (\#74150)
  \item Price, \emph{A perfect gentleman} (\#77137)
  \item Shaw, \emph{The Miraculous Revenge} (\#20336)
  \item Smith, \emph{A hat in the radio ring} (\#74149)
  \item Stuart, \emph{Don't Think About It} (\#60935)
  \item Swet, \emph{The Medici Boots} (\#32639)
  \item Terhune, \emph{The girl in the crowd} (\#69030)
  \item Towers, \emph{In case of fire} (\#79612)
  \item Tuttle, \emph{Helped by a horse doctor} (\#78539)
  \item Walton, \emph{Strange Alliance} (\#28047)
  \item Wharton, \emph{The Choice} (\#24348)
  \item Wylie, \emph{The Underpup} (\#33212)
  \item Yerxa, \emph{Phantom of the Forest} (\#32891)
\end{enumerate}
\end{multicols}

\subsection{Experiment 2 list of paper pairs}

Below is a table of the 135 OpenAI-human paper pairs used in experiment 2.
The venue listed is the venue of the human paper. 

\begingroup\small
\setlength{\tabcolsep}{4pt}
\newcolumntype{L}[1]{>{\raggedright\arraybackslash}p{#1}}
\begin{longtable}{@{}r L{0.38\linewidth} L{0.40\linewidth} L{0.15\linewidth}@{}}
\label{tab:pairs}\\
\toprule
No. & OpenAI preprint & Matched human paper & Venue \\
\midrule
\endfirsthead
\toprule
No. & OpenAI preprint & Matched human paper & Venue \\
\midrule
\endhead
\bottomrule
\endfoot
5 & A Complete Local Domain without a Small Cohen--Macaulay Module & He (2013), Normality and Cohen-Macaulayness of local models of Shimura varieties & Duke Math. J. \\
13 & A Counterexample to the Group-Ring Determinant Conjecture & Li (2012), Compact Group Automorphisms, Addition Formulas and Fuglede-Kadison Determinants & Annals of Mathematics \\
19 & A Fully Polynomial Randomized Approximation Scheme for Perfect Matchings in General Graphs & Rudelson et al. (2016), Hafnians, perfect matchings and Gaussian matrices & Ann. Probab. \\
20 & A Gepner stability condition on every smooth quintic threefold & Chiodo and Ruan (2010), Landau-Ginzburg/Calabi-Yau correspondence for quintic three-folds via symplectic transformations & Inventiones \\
22 & A Linear Clock for Random Walk on Tree-Weighted Planar Maps & Benes et al. (2013), On the rate of convergence of loop-erased random walk to SLE(2) & Comm. Math. Phys. \\
33 & A Sharp Threshold Bound for Monotone Graph Properties & Mester (2013), Invariant monotone coupling need not exist & Ann. Probab. \\
39 & A Torsion-Free Group Algebra That Is Not Directly Finite & Gardam (2021), A counterexample to the unit conjecture for group rings & Annals of Mathematics \\
53 & A counterexample to the Colin de Verdi{\`e}re chromatic conjecture & Achlioptas and Naor (2005), The two possible values of the chromatic number of a random graph & Annals of Mathematics \\
60 & A dimension-free logarithmic Sobolev inequality for subgaussian log-concave measures & Mendelson (2011), Discrepancy, chaining and subgaussian processes & Ann. Probab. \\
71 & A high-arity counterexample to Pixton completeness in Chow & Janda (2017), Relations on $\overline M_{g,n}$ via equivariant Gromov-Witten theory of $\mathbb P^1$ & Compositio \\
97 & A sharp Fourier certificate for planar circle packing & Cohn et al. (2017), The sphere packing problem in dimension 24 & Annals of Mathematics \\
101 & A smooth three-torus diffeomorphism with simple Lebesgue spectrum & Guarino et al. (2018), Rigidity of critical circle maps & Duke Math. J. \\
105 & A strict inverse-first-power bound for univalent functions & Hedenmalm and Shimorin (2005), Weighted Bergman spaces and the integral means spectrum of conformal mappings & Duke Math. J. \\
108 & A translational tile with no fully periodic tiling in dimension three & Lev and Matolcsi (2022), The Fuglede conjecture for convex domains is true in all dimensions & Acta Mathematica \\
112 & A uniform influence bound for hypergraph properties & Polcyn et al. (2020), Minimum pair degree condition for tight Hamiltonian cycles in $4$-uniform hypergraphs & Acta Math. Hungar. \\
114 & Abelian covers, Gale correspondences, and the Hodge conjecture for powers & Farb et al. (2021), The Essential Dimension of Congruence Covers & Compositio \\
118 & Abundance in numerical dimension one for terminal threefolds in positive characteristic & Xu and Zhang (2019), Nonvanishing for threefolds in characteristic $p>5$ & Duke Math. J. \\
124 & Almost-Linear-Time Maximum-Cardinality Matching in General Graphs & Bonamy et al. (2021), EPTAS and Subexponential Algorithm for Maximum Clique on Disk and Unit Ball Graphs & J. ACM \\
125 & Almost-everywhere Fourier convergence in $L\log L$ & Kowalski and Sawin (2016), Kloosterman paths and the shape of exponential sums & Compositio \\
128 & Ambiently homeomorphic isolated hypersurfaces of multiplicities two and three & Huh (2014), Milnor numbers of projective hypersurfaces with isolated singularities & Duke Math. J. \\
131 & An Endpoint Gradient Bound for the Centered Disk Maximal Operator & Melas (2003), The best constant for centered Hardy-Littlewood maximal inequality & Annals of Mathematics \\
142 & An $L^3$ bound for the trilinear Hilbert transform & Gindikin et al. (2006), Horospherical model for holomorphic discrete series and horospherical Cauchy transform & Compositio \\
161 & Anticanonical nonvanishing from smooth semipositivity & Tosatti (2026), Semipositive line bundles and (1,1)-classes & Acta Math. Sin. (Engl. Ser.) \\
166 & Arithmetic classification and non-Pisot singularity for Bernoulli convolutions & Varj{\'u} (2019), Absolute continuity of Bernoulli convolutions for algebraic parameters & J. Amer. Math. Soc. \\
170 & Average sensitivity of polynomial threshold functions & Bu et al. (2024), Complexity of quantum circuits via sensitivity, magic, and coherence & Comm. Math. Phys. \\
179 & Bloch's conjecture for surfaces with $p_g=0$ & Barbieri-Viale et al. (1996), Roitman's theorem for singular complex projective surfaces & Duke Math. J. \\
181 & Bochner--Riesz multipliers in three dimensions & Heo et al. (2011), Radial Fourier multipliers in high dimensions & Acta Mathematica \\
184 & Bounded-Step Walks on Gaussian Primes & Granville et al., When the sieve works & Duke Math. J. \\
186 & Bounded-degree coboundary expanders in every dimension & Bourgain and Varj{\'u} (2012), Expansion in $SL_d(Z/qZ)$, q arbitrary & Inventiones \\
187 & Bounded klt complements for Fano contractions & Tian and Xu (2017), Finiteness of fundamental groups & Compositio \\
194 & Canonical ampleness of compact hyperbolic K{\"a}hler manifolds & Linshaw and Song (2023), The global sections of chiral de Rham complexes on compact Ricci-flat K{\"a}hler manifolds II & Comm. Math. Phys. \\
198 & Catalan's constant is irrational & Adamczewski and Bugeaud (2007), On the complexity of algebraic number I. Expansions in integer bases & Annals of Mathematics \\
203 & Combinatorial invariance of Kazhdan--Lusztig polynomials & Fiebig (2010), The multiplicity one case of Lusztig's conjecture & Duke Math. J. \\
211 & Conditional good minimal models for compact K\"ahler fourfolds & Liu et al. (2025), Minimal model program for algebraically integrable foliations on klt varieties & Compositio \\
221 & Contact Fano manifolds and the LeBrun--Salamon conjecture & Etnyre et al. (2012), Tightness in contact metric 3-manifolds & Inventiones \\
224 & Correspondence coloring graphs with a forbidden clique & Conlon et al. (2013), Two extensions of Ramsey's theorem & Duke Math. J. \\
229 & Critical bond and site percolation on the cubic lattice & Smirnov and Schramm (2011), On the scaling limits of planar percolation & Ann. Probab. \\
231 & Critical local smoothing for the three-dimensional wave equation & Donninger (2017), Strichartz estimates in similarity coordinates and stable blowup for the critical wave equation & Duke Math. J. \\
258 & Endpoint convergence for the planar Schr\"odinger equation & Lukkarinen and Spohn (2011), Weakly nonlinear Schr{\"o}dinger equation with random initial data & Inventiones \\
259 & Endpoint pointwise convergence for the Schr\"odinger equation in higher dimensions & Goldberg and Visan (2006), A counterexample to dispersive estimates for Schr{\"o}dinger operators in higher dimensions & Comm. Math. Phys. \\
265 & Equidistribution of Prime-Degree Torus Packets with Arbitrary Local Type & Oliver et al. (2024), An approximate form of Artin's holomorphy conjecture and non-vanishing of Artin $L$-functions & Inventiones \\
266 & Equidistribution of primitive quartic torus packets for arbitrary orders & Khayutin (2019), Arithmetic of Double Torus Quotients and the Distribution of Periodic Torus Orbits & Duke Math. J. \\
267 & Equidistribution of Primitive Sextic Torus Packets & Bhargava and Harron (2016), The equidistribution of lattice shapes of rings of integers in cubic, quartic, and quintic number fields & Compositio \\
269 & Ergodicity of triangular billiards with an irrational angle & Fraczek and Ulcigrai (2014), Ergodic directions for billiards in a strip with periodically located obstacles & Comm. Math. Phys. \\
273 & Every complex K3 surface is Oka & Forstneric (2006), Runge approximation on convex sets implies the Oka property & Annals of Mathematics \\
274 & Every Four-Dimensional Kakeya Set Has Full Hausdorff Dimension & Wang and Zahl (2026), Sticky Kakeya sets and the sticky Kakeya conjecture & J. Amer. Math. Soc. \\
276 & Exact Birch--Swinnerton-Dyer Formula from Low Selmer Corank & Cesnavicius and Imai (2016), The remaining cases of the Kramer-Tunnell conjecture & Compositio \\
305 & Finite ordinary minimal model programs on compact K{\"a}hler fourfolds & Gongyo and Lehmann (2013), Reduction maps and minimal model theory & Compositio \\
310 & Fontaine--Mazur modularity at the prime 2 & Gee and Kassaei (2013), Companion Forms in Parallel Weight One & Compositio \\
314 & Frobenius Structures on Tame Hecke Eigensheaves & Chin (2004), Independence of ell of Monodromy Groups & J. Amer. Math. Soc. \\
317 & Fujita's freeness conjecture & Brown (2012), Big $q$-ample Line Bundles & Compositio \\
337 & Global quantum geometric Langlands at irrational level & Travkin (2016), Quantum geometric Langlands correspondence in positive characteristic: the GL(N) case & Duke Math. J. \\
348 & Hilbert's tenth problem over the rational numbers & Poonen (2003), Hilbert's Tenth Problem and Mazur's Conjecture for large subrings of Q & J. Amer. Math. Soc. \\
351 & Hyperbolicity Cones Without Semidefinite Lifts & Looijenga, Discrete automorphism groups of convex cones of finite type & Compositio \\
361 & Integral points on character varieties of curves & Boalch (2014), Geometry and braiding of Stokes data; Fission and wild character varieties & Annals of Mathematics \\
365 & Irrational cubic fourfolds with geometric K3 categories & Addington and Thomas (2014), Hodge theory and derived categories of cubic fourfolds & Duke Math. J. \\
367 & Koebe's Circle-Domain Conjecture & Karafyllia and Ntalampekos (2026), Uniformization of Gromov hyperbolic domains by circle domains & Duke Math. J. \\
368 & $L_1$ Embeddings of Graphs of Bounded Treewidth & Feldmann et al. (2018), A $(1 + {\varepsilon})$-Embedding of Low Highway Dimension Graphs into Bounded Treewidth Graphs & SIAM J. Comput. \\
376 & Log abundance in characteristic zero & Birkar and Hu (2014), Log canonical pairs with good augmented base loci & Compositio \\
377 & Log abundance in numerical dimension one for threefolds in positive characteristic & Fasel (2011), Stably free modules over smooth affine threefolds & Duke Math. J. \\
382 & An Upper Bound of $9/4$ for the Matrix Multiplication Exponent & Landsberg (2006), The border rank of the multiplication of two by two matrices is seven & J. Amer. Math. Soc. \\
384 & Maximal Seshadri constants on arbitrary polarized surfaces & Nystr{\"o}m (2018), Canonical growth conditions associated to ample line bundles & Duke Math. J. \\
385 & Maximal multipoint Seshadri constants in positive characteristic & Chiriv{\`i} et al. (2023), Seshadri stratifications and standard monomial theory & Inventiones \\
395 & Milne's rationality conjecture for abelian varieties & Tang (2018), Cycles in the de Rham cohomology of abelian varieties over number fields & Compositio \\
397 & Minimal metrics and interior injectivity for nef adjoints & Ambro (2014), An Injectivity Theorem & Compositio \\
400 & Modularity of elliptic curves over imaginary quadratic fields & Lei et al. (2015), Euler systems for modular forms over imaginary quadratic fields & Compositio \\
401 & Monochromatic finite sums and products in the positive integers & Croot (2003), On a coloring conjecture about unit fractions & Annals of Mathematics \\
402 & Nagata's conjecture for plane curves & Mikhalkin (2000), Real algebraic curves, the moment map and amoebas & Annals of Mathematics \\
407 & No bigeodesics in planar first-passage percolation & Damron and Hanson (2014), Busemann functions and infinite geodesics in two-dimensional first-passage percolation & Comm. Math. Phys. \\
408 & No percolation at criticality on quasi-transitive graphs & Duminil-Copin et al. (2020), Existence of phase transition for percolation using the Gaussian Free Field & Duke Math. J. \\
417 & Numerical Semiampleness of Nef Adjoint Divisors & Ejiri and Gongyo (2019), Nef anti-canonical divisors and rationally connected fibrations & Compositio \\
419 & Numerical semiampleness of nef adjoint classes on compact K{\"a}hler manifolds & Cao (2014), Numerical dimension and a Kawamata-Viehweg-Nadel type vanishing theorem on compact K{\"a}hler manifolds & Compositio \\
423 & Open Homomorphisms of Global Solvably Closed Galois Groups & Zywina (2015), The inverse Galois problem for $PSL_2(F_p)$ & Duke Math. J. \\
425 & Orbifold and logarithmic Iitaka subadditivity & Edidin et al. (2010), Logarithmic trace and orbifold products & Duke Math. J. \\
426 & Ordinary two-point correlations of multiplicative functions & Klurman (2017), Correlations of multiplicative functions and applications & Compositio \\
427 & $P=W$ in composite rank for fixed determinant & Li and Liedtke (2012), Rational Curves on K3 Surfaces & Inventiones \\
432 & Perfect completeness for 2-to-1 games & Slofstra (2020), Tsirelson's problem and an embedding theorem for groups arising from non-local games & J. Amer. Math. Soc. \\
434 & Planar Graph Metrics Embed into $L_1$ with Constant Distortion & Krauthgamer et al. (2005), Measured descent: A new embedding method for finite metrics & GAFA \\
435 & Pointwise Multiple Ergodic Averages for Mixing Transformations & Ziegler (2007), Universal Characteristic Factors and Furstenberg Averages & J. Amer. Math. Soc. \\
436 & Pointwise convergence of fourfold ergodic averages for mixing transformations & Demeter et al. (2008), Breaking the duality in the return times theorem & Duke Math. J. \\
444 & Positive Metric Entropy for the Standard Map at Large Parameters & Dobbs and Mihalache (2019), Diabolical entropy & Comm. Math. Phys. \\
450 & Positivity of Serre's Intersection Multiplicity & Dao and Takagi (2016), On the relationship between depth and cohomological dimension & Compositio \\
455 & Prime Predecessors with an Even Number of Prime Factors & Matom{\"a}ki and Shao (2017), Vinogradov's three primes theorem with almost twin primes & Compositio \\
459 & Projective Hodge lines and ordinary Iitaka subadditivity & Maulik and Poonen (2012), N{\'e}ron-Severi groups under specialization & Duke Math. J. \\
460 & Projective-space bases of Lagrangian fibrations & Rizell et al. (2016), Lagrangian isotopy of tori in $S^2 \times S^2$ and $\mathbb{C}P^2$ & GAFA \\
471 & Quasipolynomial Bounds for Arithmetic Progressions & Sanders (2011), On Roth's theorem on progressions & Annals of Mathematics \\
472 & Quaternionic division points on curves in the Siegel threefold & Levin (2016), Integral points of bounded degree on affine curves & Compositio \\
478 & Ramanujan--Arthur Decompositions of Cuspidal Functions at Full Finite Level & Lapid and Rallis (2003), On the nonnegativity of $L(\frac 12,\pi )$ for $SO(2n+1)$ & Annals of Mathematics \\
479 & Random Walks on Critical FK--Ising Maps and Liouville Brownian Motion & Garban et al. (2016), Liouville Brownian motion & Ann. Probab. \\
485 & Reconstruction from Milnor $K$-theory modulo the characteristic & Orlov et al. (2007), An exact sequence for Milnor's K-theory with applications to quadratic forms & Annals of Mathematics \\
486 & Reconstruction of Function Fields from Mod-$\ell$ Milnor $K$-Theory & Hesselholt and Madsen (2003), On the K-theory of local fields & Annals of Mathematics \\
488 & Relative denominators and effective systems for log Calabi-Yau fibrations & Whang (2020), Global geometry on moduli of local systems for surfaces with boundary & Compositio \\
491 & Removable Boundaries and Rigidity of Circle Domains & Ntalampekos and Younsi (2020), Rigidity theorems for circle domains & Inventiones \\
494 & Riesz transforms and uniform rectifiability in higher codimension & Hofmann et al. (2014), Uniform rectifiability and harmonic measure II: Poisson kernels in $L^p$ imply uniform rectifiability & Duke Math. J. \\
495 & Rigidity of Smooth Billiards with a Continuous Caustic Collar & Huang et al. (2018), On the Marked Length Spectrums of Generic Strictly Convex Billiard Tables & Duke Math. J. \\
505 & Semialgebraic universal covers of normal projective varieties & Alexeev and Pardini (2012), Non-normal abelian covers & Compositio \\
536 & Squarefree values of quartics and power-free values of polynomials & Poonen (2003), Squarefree values of multivariable polynomials & Duke Math. J. \\
553 & Symplectic Balls in Symmetric Polar Products & Pinsonnault (2008), Symplectomorphism groups and embeddings of balls into rational ruled 4-manifolds & Compositio \\
554 & Talagrand's discrete-convexity conjecture & Gozlan et al. (2011), A new characterization of Talagrand's transport-entropy inequalities and applications & Ann. Probab. \\
555 & Tame Hecke Eigensheaves with Several Marked Points & Nadler and Yun (2019), Geometric Langlands correspondence for SL(2), PGL(2) over the pair of pants & Compositio \\
559 & Temperedness at ramified places for globally generic exceptional groups & Jiang and Soudry (2003), The local converse theorem for SO(2n+1) and applications & Annals of Mathematics \\
561 & Termination for projective log canonical fourfolds with rational boundary & Birkar (2010), On existence of log minimal models & Compositio \\
565 & Termination of generalized log canonical flips on compact K{\"a}hler fourfolds & Chen and Tsakanikas (2023), On the termination of flips for log canonical generalized pairs & Acta Math. Sin. (Engl. Ser.) \\
566 & The abelian Zilber--Pink conjecture & Poonen and Voloch (2010), The Brauer-Manin obstruction for subvarieties of abelian varieties over function fields & Annals of Mathematics \\
570 & The Blockwise Alperin Weight Conjecture & Isaacs and Navarro (2002), New refinements of the McKay conjecture for arbitrary finite groups & Annals of Mathematics \\
581 & The Falconer distance conjecture in all dimensions & Filho and Vallentin (2010), Fourier analysis, linear programming, and densities of distance avoiding sets in $R^n$ & J. Eur. Math. Soc. \\
590 & The Kakeya maximal conjecture in three dimensions & Wang and Zahl (2025), The Assouad dimension of Kakeya sets in $\mathbb{R}^3$ & Inventiones \\
595 & The Mahler Conjecture for General Convex Bodies & Artstein-Avidan et al., From Symplectic Measurements to the Mahler Conjecture & Duke Math. J. \\
602 & The Poisson--Dirichlet law for prime predecessors & Diaconis et al. (2004), The Poisson-Dirichlet law is the unique invariant distribution for uniform split-merge transformations & Ann. Probab. \\
606 & The Restricted Geometric Langlands Equivalence in Positive Characteristic & Chen and Zhu (2017), Geometric Langlands in prime characteristic & Compositio \\
607 & The Selmer converse for elliptic curves at every prime & Klagsbrun et al. (2013), Disparity in Selmer ranks of quadratic twists of elliptic curves & Annals of Mathematics \\
614 & The strong hyperk\"ahler SYZ conjecture & Verbitsky (2010), Hyperkahler SYZ conjecture and semipositive line bundles & GAFA \\
625 & The critical Liouville quantum sphere and geometric limits of FK maps at $q=4$ & Ding and Dunlap (2020), Subsequential scaling limits for Liouville graph distance & Comm. Math. Phys. \\
629 & The crossing number of complete bipartite graphs & Liu and Zhang (2023), Maximizing the Minimum and Maximum Forcing Numbers of Perfect Matchings of Graphs & Acta Math. Sin. (Engl. Ser.) \\
630 & The crossing number of complete graphs & Lara and Rubio-Montiel (2019), On crossing families of complete geometric graphs & Acta Math. Hungar. \\
634 & The entropy-rate dimension formula for self-similar measures on the line & Neunh{\"a}userer (2006), A construction of singular overlapping asymmetric self-similar measures & Acta Math. Hungar. \\
635 & The equivariant cohomological Hikita conjecture for arbitrary quivers & Davison et al. (2015), Purity for graded potentials and quantum cluster positivity & Compositio \\
645 & The irrationality exponent of $\pi$ is $2$ & Zudilin (2014), Two hypergeometric tales and a new irrationality measure of $\zeta (2)$ & Annals of Mathematics \\
650 & The logarithmic Brunn--Minkowski conjecture & Schuster (2010), Crofton measures and Minkowski valuations & Duke Math. J. \\
653 & The maximal triangular Hilbert transform at the symmetric point & Ban and Schlichtkrull (2006), A Paley-Wiener theorem for reductive symmetric spaces & Annals of Mathematics \\
655 & The mean analytic rank of quadratic twists of elliptic curves & Bhargava et al. (2019), Three-isogeny Selmer groups and ranks of abelian varieties in quadratic twist families over a number field & Duke Math. J. \\
661 & The p-adic section conjecture & Parimala and Suresh (2010), The u-invariant of the function fields of p-adic curves & Annals of Mathematics \\
664 & The rational Hodge conjecture for CM abelian varieties & Milne (2002), Polarizations and Grothendieck's Standard Conjectures & Annals of Mathematics \\
676 & The strong thin tree conjecture & Pemantle (1991), Choosing a Spanning Tree for the Integer Lattice Uniformly & Ann. Probab. \\
677 & The symmetric Mahler conjecture and its equality cases & Iriyeh and Shibata (2020), Symmetric Mahler's conjecture for the volume product in the three dimensional case & Duke Math. J. \\
679 & The two-primary Birch--Swinnerton-Dyer formula in Selmer corank at most one & Taelman (2012), Special L-values of Drinfeld modules & Annals of Mathematics \\
691 & Tree Constructions for the $\ell_1$ Distortion of Binary Edit Distance & Pardon (2011), On the distortion of knots on embedded surfaces & Annals of Mathematics \\
695 & Two-step monodromy of special quasi-projective varieties & Taji (2016), The isotriviality of families of canonically-polarized manifolds over a special quasi-projective base & Compositio \\
700 & Uniform Cartier sections for Fano type contractions & Blum et al. (2023), The existence of the K{\"a}hler-Ricci soliton degeneration & Forum Math. Pi \\
704 & Uniform Pluricanonical Iitaka Fibrations & Birkar and Zhang (2016), Effectivity of Iitaka fibrations and pluricanonical systems of polarized pairs & Publ. Math. IHES \\
721 & Universal optimality of the triangular lattice & Cohn and Kumar (2007), Universally optimal distribution of points on spheres & J. Amer. Math. Soc. \\
731 & Virasoro Constraints under Projectivization & Adler and Moerbeke (1995), Matrix Integrals, Toda symmetries, Virasoro constraints, and orthogonal polynomials & Duke Math. J. \\
738 & Weil classes and Hodge classes on abelian powers & Abdulali (2002), Hodge structures on abelian varieties of type III & Annals of Mathematics \\
743 & Ample rank-two bundles on the quadric surface without Griffiths-positive metrics & Schreieder (2019), On the rationality problem for quadric bundles & Duke Math. J. \\
746 & The additive indecomposability of the primes & Shao (2014), A Density Version of the Vinogradov Three Primes Theorem & Duke Math. J. \\
\end{longtable}
\endgroup

\subsection{Experiment 3 list of models and scores}

Below is a list of all models used for experiment 3, along with each Model's ECI score, median total persistence / null, as well as AIME accuracy on the 30 problems.
The list of problems can be found at \cite{matharena_aime2026}.

\begingroup\small
\setlength{\tabcolsep}{6pt}
\begin{longtable}{@{}l l c c c@{}}
\label{tab:eci-models}\\
\toprule
Model & Lab & ECI & Total persistence / null & AIME accuracy \\
\midrule
\endfirsthead
\toprule
Model & Lab & ECI & Total persistence / null & AIME accuracy \\
\midrule
\endhead
\bottomrule
\endfoot
Kimi K3 & Moonshot (Kimi) & 157.4 & 0.492 & 28/30 \\
GLM-5.3 & Zhipu (GLM) & 155.6 & 0.348 & 23/30 \\
DeepSeek V4 Pro 0813 & DeepSeek & 155.3 & 0.344 & 29/30 \\
Qwen3.8 Max (0902) & Qwen & 155.1 & 0.428 & 29/30 \\
DeepSeek V4.1 Flash & DeepSeek & 154.9 & 0.368 & 29/30 \\
DeepSeek V4 Flash 0731 & DeepSeek & 154.5 & 0.369 & 28/30 \\
Qwen3.7-Max & Qwen & 153.7 & 0.369 & 28/30 \\
GLM-5.3-Flash & Zhipu (GLM) & 151.9 & 0.390 & 19/30 \\
GLM-5.2 & Zhipu (GLM) & 151.8 & 0.403 & 26/30 \\
Kimi K2.6 & Moonshot (Kimi) & 151.1 & 0.395 & 28/30 \\
Inkling-Small & Thinking Machines & 150.2 & 0.395 & 29/30 \\
Kimi K2.7 Code & Moonshot (Kimi) & 150.0 & 0.454 & 29/30 \\
GLM-5.1 & Zhipu (GLM) & 149.8 & 0.322 & 26/30 \\
Qwen 3.8 27B & Qwen & 149.4 & 0.401 & 28/30 \\
DeepSeek-V4-Pro & DeepSeek & 149.1 & 0.391 & 29/30 \\
Inkling & Thinking Machines & 148.5 & 0.356 & 28/30 \\
Kimi K2.5 & Moonshot (Kimi) & 148.0 & 0.370 & 27/30 \\
Qwen 3.6 Plus & Qwen & 147.7 & 0.405 & 29/30 \\
Qwen3.7-Plus & Qwen & 147.4 & 0.381 & 29/30 \\
MiniMax-M3 & MiniMax & 146.9 & 0.368 & 23/30 \\
MiniMax-M2.5 & MiniMax & 146.7 & 0.371 & 28/30 \\
Qwen3.5 397B-A17B & Qwen & 146.7 & 0.419 & 28/30 \\
Qwen3.6 27B & Qwen & 146.5 & 0.448 & 27/30 \\
DeepSeek-V3.2 & DeepSeek & 146.3 & 0.408 & 28/30 \\
Nemotron 3 Ultra & NVIDIA & 146.2 & 0.347 & 26/30 \\
DeepSeek-V4-Flash & DeepSeek & 146.1 & 0.305 & 27/30 \\
Kimi K2 Thinking & Moonshot (Kimi) & 146.0 & 0.377 & 26/30 \\
MiniMax-M2.7 & MiniMax & 145.8 & 0.332 & 27/30 \\
GLM-5 & Zhipu (GLM) & 145.8 & 0.402 & 27/30 \\
DeepSeek-V3.2-Exp & DeepSeek & 145.0 & 0.392 & 28/30 \\
Qwen3.7 Flash & Qwen & 144.6 & 0.515 & 26/30 \\
Qwen 3.6 35B-A3B & Qwen & 143.9 & 0.486 & 27/30 \\
Qwen3-235B-A22B-Thinking (Jul 2025) & Qwen & 143.8 & 0.428 & 18/30 \\
GLM-4.7 & Zhipu (GLM) & 143.5 & 0.389 & 27/30 \\
Qwen 3.6 Flash & Qwen & 143.3 & 0.447 & 28/30 \\
Gemma 4 31B IT & Google & 142.7 & 0.452 & 27/30 \\
Qwen3.5-35B-A3B & Qwen & 142.5 & 0.428 & 27/30 \\
Gemma 4 26B A4B & Google & 141.8 & 0.444 & 25/30 \\
Mistral Medium 3.5 & Mistral & 141.3 & 0.426 & 27/30 \\
DeepSeek-R1 (May 2025) & DeepSeek & 141.3 & 0.400 & 27/30 \\
gpt-oss-120b & OpenAI & 139.9 & 0.430 & 27/30 \\
DeepSeek-V3.1 & DeepSeek & 139.9 & 0.398 & 9/30 \\
Qwen3.5-9B & Qwen & 139.5 & 0.456 & 22/30 \\
DeepSeek-R1 & DeepSeek & 139.0 & 0.403 & 14/30 \\
Qwen3-32B & Qwen & 138.5 & 0.515 & 9/30 \\
Qwen3-14B & Qwen & 138.2 & 0.494 & 21/30 \\
gpt-oss-20b & OpenAI & 137.8 & 0.390 & 24/30 \\
Qwen3-30B-A3B & Qwen & 136.2 & 0.432 & 16/30 \\
\end{longtable}
\endgroup

%% file: main.bbl
\newcommand{\etalchar}[1]{$^{#1}$}
\providecommand{\bysame}{\leavevmode\hbox to3em{\hrulefill}\thinspace}
\providecommand{\MR}{\relax\ifhmode\unskip\space\fi MR }
\providecommand{\MRhref}[2]{%
  \href{http://www.ams.org/mathscinet-getitem?mr=#1}{#2}
}
\providecommand{\href}[2]{#2}
\begin{thebibliography}{FGRW{\etalchar{+}}26}

\bibitem[Bau21]{bauer:ripser}
Ulrich Bauer, \emph{Ripser: efficient computation of {V}ietoris-{R}ips
  persistence barcodes}, J. Appl. Comput. Topol. \textbf{5} (2021), no.~3,
  391--423, \href {http://dx.doi.org/10.1007/s41468-021-00071-5}
  {\path{doi:10.1007/s41468-021-00071-5}},
  \url{https://doi.org/10.1007/s41468-021-00071-5}.

\bibitem[BHG{\etalchar{+}}25]{baker2025monitoring}
Bowen Baker, Joost Huizinga, Leo Gao, Zehao Dou, Melody~Y. Guan, Aleksander
  Madry, Wojciech Zaremba, Jakub Pachocki, and David Farhi, \emph{Monitoring
  reasoning models for misbehavior and the risks of promoting obfuscation},
  2025, \href {http://arxiv.org/abs/2503.11926} {\path{arXiv:2503.11926}},
  \href {http://dx.doi.org/10.48550/arXiv.2503.11926}
  {\path{doi:10.48550/arXiv.2503.11926}},
  \url{https://arxiv.org/abs/2503.11926}.

\bibitem[BKR14]{clear-and-compress}
Ulrich Bauer, Michael Kerber, and Jan Reininghaus, \emph{Clear and compress:
  Computing persistent homology in chunks}, Topological Methods in Data
  Analysis and Visualization III (Cham) (Peer-Timo Bremer, Ingrid Hotz, Valerio
  Pascucci, and Ronald Peikert, eds.), Springer International Publishing, 2014,
  pp.~103--117.

\bibitem[BM12]{barmak-minian}
Jonathan~Ariel Barmak and Elias~Gabriel Minian, \emph{Strong homotopy types,
  nerves and collapses}, Discrete Comput. Geom. \textbf{47} (2012), no.~2,
  301--328, \href {http://dx.doi.org/10.1007/s00454-011-9357-5}
  {\path{doi:10.1007/s00454-011-9357-5}},
  \url{https://doi.org/10.1007/s00454-011-9357-5}.

\bibitem[BP20]{Boissonnat-Pritam-edge}
Jean-Daniel Boissonnat and Siddharth Pritam, \emph{Edge collapse and
  persistence of flag complexes}, 36th {I}nternational {S}ymposium on
  {C}omputational {G}eometry, LIPIcs. Leibniz Int. Proc. Inform., vol. 164,
  Schloss Dagstuhl. Leibniz-Zent. Inform., Wadern, 2020, pp.~Art. No. 19, 15.

\bibitem[CSBB20]{christianson2020architecture}
Nicolas~H. Christianson, Ann Sizemore~Blevins, and Danielle~S. Bassett,
  \emph{Architecture and evolution of semantic networks in mathematics texts},
  Proceedings of the Royal Society A: Mathematical, Physical and Engineering
  Sciences \textbf{476} (2020), no.~2239, 20190741, \href
  {http://dx.doi.org/10.1098/rspa.2019.0741}
  {\path{doi:10.1098/rspa.2019.0741}},
  \url{https://doi.org/10.1098/rspa.2019.0741}.

\bibitem[{Epo}25]{EpochEci2025}
{Epoch AI}, \emph{{Epoch Capabilities Index}}, October 2025,
  \url{https://epoch.ai/eci}.

\bibitem[FGRW{\etalchar{+}}26]{fay2026shape}
Aideen Fay, In{\'e}s Garc{\'i}a-Redondo, Qiquan Wang, Haim Dubossarsky, and
  Anthea Monod, \emph{The shape of adversarial influence: Characterizing {LLM}
  latent spaces with persistent homology}, 2026, Version 3, 24 April 2026;
  first posted in 2025, \href {http://arxiv.org/abs/2505.20435}
  {\path{arXiv:2505.20435}}, \href
  {http://dx.doi.org/10.48550/arXiv.2505.20435}
  {\path{doi:10.48550/arXiv.2505.20435}},
  \url{https://arxiv.org/abs/2505.20435v3}.

\bibitem[For98]{forman-morse}
Robin Forman, \emph{Morse theory for cell complexes}, Adv. Math. \textbf{134}
  (1998), no.~1, 90--145, \href {http://dx.doi.org/10.1006/aima.1997.1650}
  {\path{doi:10.1006/aima.1997.1650}},
  \url{https://doi.org/10.1006/aima.1997.1650}.

\bibitem[Gab72]{gabriel:decomposition}
Peter Gabriel, \emph{Unzerlegbare {D}arstellungen. {I}}, Manuscripta Math.
  \textbf{6} (1972), 71--103; correction, ibid. 6 (1972), 309, \href
  {http://dx.doi.org/10.1007/BF01298413} {\path{doi:10.1007/BF01298413}},
  \url{https://doi.org/10.1007/BF01298413}.

\bibitem[GCP{\etalchar{+}}23]{golovneva2023roscoe}
Olga Golovneva, Moya Chen, Spencer Poff, Martin Corredor, Luke Zettlemoyer,
  Maryam Fazel-Zarandi, and Asli Celikyilmaz, \emph{{ROSCOE}: A suite of
  metrics for scoring step-by-step reasoning}, The Eleventh International
  Conference on Learning Representations, 2023, \href
  {http://arxiv.org/abs/2212.07919} {\path{arXiv:2212.07919}},
  \url{https://openreview.net/forum?id=xYlJRpzZtsY}.

\bibitem[GVP{\etalchar{+}}25]{gardinazzi2025persistent}
Yuri Gardinazzi, Karthik Viswanathan, Giada Panerai, Alessio Ansuini, Alberto
  Cazzaniga, and Matteo Biagetti, \emph{Persistent topological features in
  large language models}, Proceedings of the 42nd International Conference on
  Machine Learning, Proceedings of Machine Learning Research, vol. 267, PMLR,
  2025, pp.~18811--18830,
  \url{https://proceedings.mlr.press/v267/gardinazzi25a.html}.

\bibitem[GWC{\etalchar{+}}26]{guan2026monitoring}
Melody~Y. Guan, Miles Wang, Micah Carroll, Zehao Dou, Annie~Y. Wei, Marcus
  Williams, Benjamin Arnav, Joost Huizinga, Ian~D. Kivlichan, Amelia Glaese,
  Jakub Pachocki, and Bowen Baker, \emph{Monitoring monitorability},
  Proceedings of the 43rd International Conference on Machine Learning,
  Proceedings of Machine Learning Research, vol. 306, PMLR, 2026,
  pp.~37538--37602, \url{https://proceedings.mlr.press/v306/guan26h.html}.

\bibitem[HDA{\etalchar{+}}25]{ho2025rosetta}
Anson Ho, Jean-Stanislas Denain, David Atanasov, Samuel Albanie, and Rohin
  Shah, \emph{A rosetta stone for {AI} benchmarks}, 2025, \href
  {http://arxiv.org/abs/2512.00193} {\path{arXiv:2512.00193}}, \href
  {http://dx.doi.org/10.48550/arXiv.2512.00193}
  {\path{doi:10.48550/arXiv.2512.00193}},
  \url{https://arxiv.org/abs/2512.00193}.

\bibitem[KBB{\etalchar{+}}25]{korbak2025monitorability}
Tomek Korbak, Mikita Balesni, Elizabeth Barnes, Yoshua Bengio, Joe Benton,
  Joseph Bloom, Mark Chen, Alan Cooney, Allan Dafoe, Anca Dragan, Scott Emmons,
  Owain Evans, David Farhi, Ryan Greenblatt, Dan Hendrycks, Marius Hobbhahn,
  Evan Hubinger, Geoffrey Irving, Erik Jenner, Daniel Kokotajlo, Victoria
  Krakovna, Shane Legg, David Lindner, David Luan, Aleksander M{\k{a}}dry,
  Julian Michael, Neel Nanda, Dave Orr, Jakub Pachocki, Ethan Perez, Mary
  Phuong, Fabien Roger, Joshua Saxe, Buck Shlegeris, Mart{\'i}n Soto, Eric
  Steinberger, Jasmine Wang, Wojciech Zaremba, Bowen Baker, Rohin Shah, and
  Vlad Mikulik, \emph{Chain of thought monitorability: A new and fragile
  opportunity for {AI} safety}, 2025, \href {http://arxiv.org/abs/2507.11473}
  {\path{arXiv:2507.11473}}, \href
  {http://dx.doi.org/10.48550/arXiv.2507.11473}
  {\path{doi:10.48550/arXiv.2507.11473}},
  \url{https://arxiv.org/abs/2507.11473}.

\bibitem[KGR{\etalchar{+}}22]{kojima2022zeroshot}
Takeshi Kojima, Shixiang~Shane Gu, Machel Reid, Yutaka Matsuo, and Yusuke
  Iwasawa, \emph{Large language models are zero-shot reasoners}, Advances in
  Neural Information Processing Systems, vol.~35, 2022, \href
  {http://dx.doi.org/10.52202/068431-1613} {\path{doi:10.52202/068431-1613}},
  \url{https://proceedings.neurips.cc/paper/2022/hash/8bb0d291acd4acf06ef112099c16f326-Abstract-Conference.html}.

\bibitem[KSC{\etalchar{+}}25]{kutasov2025shade}
Jonathan Kutasov, Yuqi Sun, Paul Colognese, Teun van~der Weij, Linda Petrini,
  Chen Bo~Calvin Zhang, John Hughes, Xiang Deng, Henry Sleight, Tyler Tracy,
  Buck Shlegeris, and Joe Benton, \emph{{SHADE-Arena}: Evaluating sabotage and
  monitoring in {LLM} agents}, 2025, \href {http://arxiv.org/abs/2506.15740}
  {\path{arXiv:2506.15740}}, \href
  {http://dx.doi.org/10.48550/arXiv.2506.15740}
  {\path{doi:10.48550/arXiv.2506.15740}},
  \url{https://arxiv.org/abs/2506.15740}.

\bibitem[LKB{\etalchar{+}}24]{lightman2024verify}
Hunter Lightman, Vineet Kosaraju, Yura Burda, Harri Edwards, Bowen Baker, Teddy
  Lee, Jan Leike, John Schulman, Ilya Sutskever, and Karl Cobbe, \emph{Let's
  verify step by step}, The Twelfth International Conference on Learning
  Representations, 2024,
  \url{https://proceedings.iclr.cc/paper_files/paper/2024/hash/aca97732e30bcf1303bc22ac3924fd16-Abstract-Conference.html}.

\bibitem[LZL{\etalchar{+}}26]{li2026understanding}
Chenghao Li, Chaoning Zhang, Yi~Lu, Shuxu Chen, Xudong Wang, Jiaquan Zhang,
  Zhicheng Wang, Zhengxun Jin, Kuien Liu, Sung-Ho Bae, Guoqing Wang, Yang Yang,
  and Heng~Tao Shen, \emph{Understanding chain-of-thought in large language
  models via topological data analysis}, 2026, Version 2, 13 February 2026;
  first posted in 2025, \href {http://arxiv.org/abs/2512.19135}
  {\path{arXiv:2512.19135}}, \href
  {http://dx.doi.org/10.48550/arXiv.2512.19135}
  {\path{doi:10.48550/arXiv.2512.19135}},
  \url{https://arxiv.org/abs/2512.19135v2}.

\bibitem[{Mat}26]{matharena_aime2026}
{MathArena}, \emph{{AIME 2026}}, Hugging Face dataset
  \texttt{MathArena/aime\_2026},
  \url{https://huggingface.co/datasets/MathArena/aime_2026}, 2026, Licence CC
  BY-NC-SA 4.0.

\bibitem[MN13]{mischaikow-nanda-morse}
Konstantin Mischaikow and Vidit Nanda, \emph{Morse theory for filtrations and
  efficient computation of persistent homology}, Discrete Comput. Geom.
  \textbf{50} (2013), no.~2, 330--353, \href
  {http://dx.doi.org/10.1007/s00454-013-9529-6}
  {\path{doi:10.1007/s00454-013-9529-6}},
  \url{https://doi.org/10.1007/s00454-013-9529-6}.

\bibitem[MZB{\etalchar{+}}25]{more2025optimizing}
Abhishek More, Anthony Zhang, Nicole Bonilla, Ashvik Vivekan, Kevin Zhu, Parham
  Sharafoleslami, and Maheep Chaudhary, \emph{Optimizing chain-of-thought
  confidence via topological and {Dirichlet} risk analysis}, 2025, \href
  {http://arxiv.org/abs/2511.06437} {\path{arXiv:2511.06437}}, \href
  {http://dx.doi.org/10.48550/arXiv.2511.06437}
  {\path{doi:10.48550/arXiv.2511.06437}},
  \url{https://arxiv.org/abs/2511.06437}.

\bibitem[{Ope}26]{openai2026math}
{OpenAI}, \emph{Sharing {AI} progress in mathematics}, OpenAI research
  announcement, October 2026, Published 6 October 2026.,
  \url{https://openai.com/index/sharing-ai-progress-in-mathematics/}.

\bibitem[{Pro}]{projectgutenberg}
{Project Gutenberg}, \emph{{Project Gutenberg}},
  \url{https://www.gutenberg.org/}.

\bibitem[PSZB23]{prasad2023receval}
Archiki Prasad, Swarnadeep Saha, Xiang Zhou, and Mohit Bansal, \emph{{ReCEval}:
  Evaluating reasoning chains via correctness and informativeness}, Proceedings
  of the 2023 Conference on Empirical Methods in Natural Language Processing,
  Association for Computational Linguistics, 2023, pp.~10066--10086, \href
  {http://dx.doi.org/10.18653/v1/2023.emnlp-main.622}
  {\path{doi:10.18653/v1/2023.emnlp-main.622}},
  \url{https://aclanthology.org/2023.emnlp-main.622/}.

\bibitem[PXP26]{pan2026topology}
Yurui Pan, Ke~Xu, and Bo~Peng, \emph{Topology-enhanced alignment for large
  language models: Trajectory topology loss and topological preference
  optimization}, Findings of the Association for Computational Linguistics: ACL
  2026, Association for Computational Linguistics, 2026, pp.~24807--24821,
  \href {http://dx.doi.org/10.18653/v1/2026.findings-acl.1242}
  {\path{doi:10.18653/v1/2026.findings-acl.1242}},
  \url{https://aclanthology.org/2026.findings-acl.1242/}.

\bibitem[{Ran}]{ranks_stopwords}
{Ranks NL}, \emph{Stopwords}, \url{https://www.ranks.nl/stopwords}.

\bibitem[RHvN{\etalchar{+}}24]{ruppik2024local}
Benjamin~Matthias Ruppik, Michael Heck, Carel van Niekerk, Renato Vukovic,
  Hsien-chin Lin, Shutong Feng, Marcus Zibrowius, and Milica Ga{\v{s}}i{\'c},
  \emph{Local topology measures of contextual language model latent spaces with
  applications to dialogue term extraction}, Proceedings of the 25th Annual
  Meeting of the Special Interest Group on Discourse and Dialogue, Association
  for Computational Linguistics, 2024, pp.~344--356, \href
  {http://dx.doi.org/10.18653/v1/2024.sigdial-1.31}
  {\path{doi:10.18653/v1/2024.sigdial-1.31}},
  \url{https://aclanthology.org/2024.sigdial-1.31/}.

\bibitem[SCLJ18]{salnikov2018cooccurrence}
Vsevolod Salnikov, Daniele Cassese, Renaud Lambiotte, and Nick~S. Jones,
  \emph{Co-occurrence simplicial complexes in mathematics: identifying the
  holes of knowledge}, Applied Network Science \textbf{3} (2018), 37, \href
  {http://dx.doi.org/10.1007/s41109-018-0074-3}
  {\path{doi:10.1007/s41109-018-0074-3}},
  \url{https://link.springer.com/article/10.1007/s41109-018-0074-3}.

\bibitem[TLTK26]{tan2026shape}
Xue~Wen Tan, Galen Lee, Nathaniel Tan, and Stanley Kok, \emph{The shape of
  reasoning: Topological analysis of reasoning traces in large language
  models}, 2026, Version 3, 27 May 2026; first posted in 2025. Accepted at the
  ICML 2026 Workshop on Epistemic Intelligence in Machine Learning, \href
  {http://arxiv.org/abs/2510.20665} {\path{arXiv:2510.20665}}, \href
  {http://dx.doi.org/10.48550/arXiv.2510.20665}
  {\path{doi:10.48550/arXiv.2510.20665}},
  \url{https://arxiv.org/abs/2510.20665v3}.

\bibitem[WWS{\etalchar{+}}22]{wei2022chain}
Jason Wei, Xuezhi Wang, Dale Schuurmans, Maarten Bosma, Brian Ichter, Fei Xia,
  Ed~H. Chi, Quoc~V. Le, and Denny Zhou, \emph{Chain-of-thought prompting
  elicits reasoning in large language models}, Advances in Neural Information
  Processing Systems, vol.~35, 2022, \href
  {http://dx.doi.org/10.52202/068431-1800} {\path{doi:10.52202/068431-1800}},
  \url{https://proceedings.neurips.cc/paper/2022/hash/9d5609613524ecf4f15af0f7b31abca4-Abstract.html}.

\bibitem[ZSZ{\etalchar{+}}26]{zhang2026tdarc}
Jiaquan Zhang, Qigan Sun, Chaoning Zhang, Xudong Wang, Zhenzhen Huang, Yitian
  Zhou, Pengcheng Zheng, Chi-lok~Andy Tai, Sung-Ho Bae, Zeyu Ma, Caiyan Qin,
  Jinyu Guo, Yang Yang, and Hengtao Shen, \emph{{TDA-RC}: Task-driven alignment
  for knowledge-based reasoning chains in large language models}, 2026, \href
  {http://arxiv.org/abs/2604.04942} {\path{arXiv:2604.04942}}, \href
  {http://dx.doi.org/10.48550/arXiv.2604.04942}
  {\path{doi:10.48550/arXiv.2604.04942}},
  \url{https://arxiv.org/abs/2604.04942}.

\bibitem[ZZC{\etalchar{+}}26]{zhang2026global}
Jiaquan Zhang, Chaoning Zhang, Shuxu Chen, Xudong Wang, Zhenzhen Huang,
  Pengcheng Zheng, Shuai Yuan, Sheng Zheng, Qigan Sun, Jie Zou, Lik-Hang Lee,
  and Yang Yang, \emph{Learning global hypothesis space for enhancing
  synergistic reasoning chain}, International Conference on Learning
  Representations (ICLR), 2026,
  \url{https://proceedings.iclr.cc/paper_files/paper/2026/hash/944b457bd8e4cbfff9de9d319dde5386-Abstract-Conference.html}.

\bibitem[ZZZ{\etalchar{+}}25]{zheng2025processbench}
Chujie Zheng, Zhenru Zhang, Beichen Zhang, Runji Lin, Keming Lu, Bowen Yu,
  Dayiheng Liu, Jingren Zhou, and Junyang Lin, \emph{{ProcessBench}:
  Identifying process errors in mathematical reasoning}, Proceedings of the
  63rd Annual Meeting of the Association for Computational Linguistics (Volume
  1: Long Papers), Association for Computational Linguistics, 2025,
  pp.~1009--1024, \href {http://dx.doi.org/10.18653/v1/2025.acl-long.50}
  {\path{doi:10.18653/v1/2025.acl-long.50}},
  \url{https://aclanthology.org/2025.acl-long.50/}.

\end{thebibliography}
